\documentclass{article}

\usepackage{PRIMEarxiv}

\usepackage[utf8]{inputenc}
\usepackage[T1]{fontenc}
\usepackage{hyperref}
\hypersetup{
  pdftitle={Learning General Representation of 12-Lead ECG With a Joint-Embedding Predictive Architecture},
  pdfauthor={Sehun Kim},
  pdfsubject={Self-supervised learning for 12-lead ECG representation learning},
  pdfkeywords={ECG, self-supervised learning, representation learning, transformers}
}
\usepackage{url}
\usepackage{graphicx}
\graphicspath{{pics/}}
\usepackage{subcaption}
\usepackage{multirow}
\usepackage{threeparttable}
\usepackage{array}
\usepackage{wrapfig}
\usepackage{booktabs}
\usepackage{adjustbox}
\usepackage{xcolor}
\usepackage{amsmath}
\usepackage{amssymb}
\usepackage[numbers]{natbib}
\usepackage{caption}
\usepackage{microtype}
\usepackage{fancyhdr}

\title{Learning General Representation of 12-Lead ECG\\ With a Joint-Embedding Predictive Architecture}

\author{
  Sehun Kim\thanks{Corresponding author: \texttt{shunhun33@gmail.com}} \\
  Samsung Medical Center \\
  81 Irwon-ro, Gangnam-gu, Seoul 06351, South Korea \\
  \texttt{shunhun33@gmail.com}
}

\begin{document}
\maketitle

\begin{abstract}
Electrocardiogram (ECG) captures the heart's electrical signals, offering valuable information for diagnosing cardiac conditions. However, the scarcity of labeled data makes it challenging to fully leverage supervised learning in the medical domain. Self-supervised learning (SSL) offers a promising solution, enabling models to learn from unlabeled data and uncover meaningful patterns. In this paper, we show that masked modeling in the latent space can be a powerful alternative to existing self-supervised methods in the ECG domain. We introduce ECG-JEPA, an SSL model for 12-lead ECG analysis that learns semantic representations of ECG data by predicting in the hidden latent space, bypassing the need to reconstruct raw signals. This approach offers several advantages in the ECG domain: (1) it avoids producing unnecessary details, such as noise, which is common in ECG; and (2) it addresses the limitations of naïve L2 loss between raw signals. Another key contribution is the introduction of Cross-Pattern Attention (CroPA), a specialized masked attention mechanism tailored for 12-lead ECG data. ECG-JEPA is trained on the union of several open ECG datasets, totaling approximately 180,000 samples, and achieves state-of-the-art performance in various downstream tasks including diagnostic classification, feature extraction, and segmentation.
\end{abstract}


\section{Introduction}
Electrocardiography is a non-invasive method to measure the electrical activity of the heart over time, serving as a crucial tool for diagnosing various cardiac conditions. While numerous supervised methods have been developed to detect heart diseases using ECG data \citep{ hannun2019cardiologist, ribeiro2020automatic, siontis2021artificial}, these models often face significant performance degradation when applied to data distributions different from those on which they were trained. This challenge points to the need for more flexible approaches that can learn robust, transferable representations from ECG data.

Self-supervised learning (SSL) offers an alternative approach by learning general representations in diverse domains, such as natural language processing (NLP) \citep{bert, gpt3, llama}, computer vision (CV) \citep{simclr, mae, ijepa}, and video analysis \citep{vmae, vjepa}. Despite this promise, the application of SSL to ECG data presents unique challenges. For instance, data augmentation, which is essential in many SSL architectures, is more complex for ECG than for computer vision data. Simple transformations like rotation, scaling, and flipping, effective in CV, can distort the physiological meaning of ECG signals. Additionally, ECG recordings often contain artifacts and noise, which may cause autoencoder-based SSL models to struggle with reconstructing raw signals. These architectures may also miss visually subtle but diagnostically critical features, such as P-waves and T-waves, which are imperative for diagnosing certain cardiac conditions.

In this work, we propose ECG Joint-Embedding Predictive Architecture (ECG-JEPA) tailored for 12-lead ECG data, effectively addressing the aforementioned challenges. ECG-JEPA utilizes a transformer architecture to capture the semantic meaning of the ECG. By masking several patches of the ECG, ECG-JEPA predicts abstract representations of the missing segments, indicating a high-level understanding of the data. Additionally, we develop a novel masked-attention for multi-lead ECG data, which we call Cross-Pattern Attention (CroPA). CroPA incorporates clinical knowledge into the model as an inductive bias, guiding it to focus on clinically relevant patterns and relationships across leads.

\begin{figure*}[t!]
    \centering
    \includegraphics[width=\textwidth]{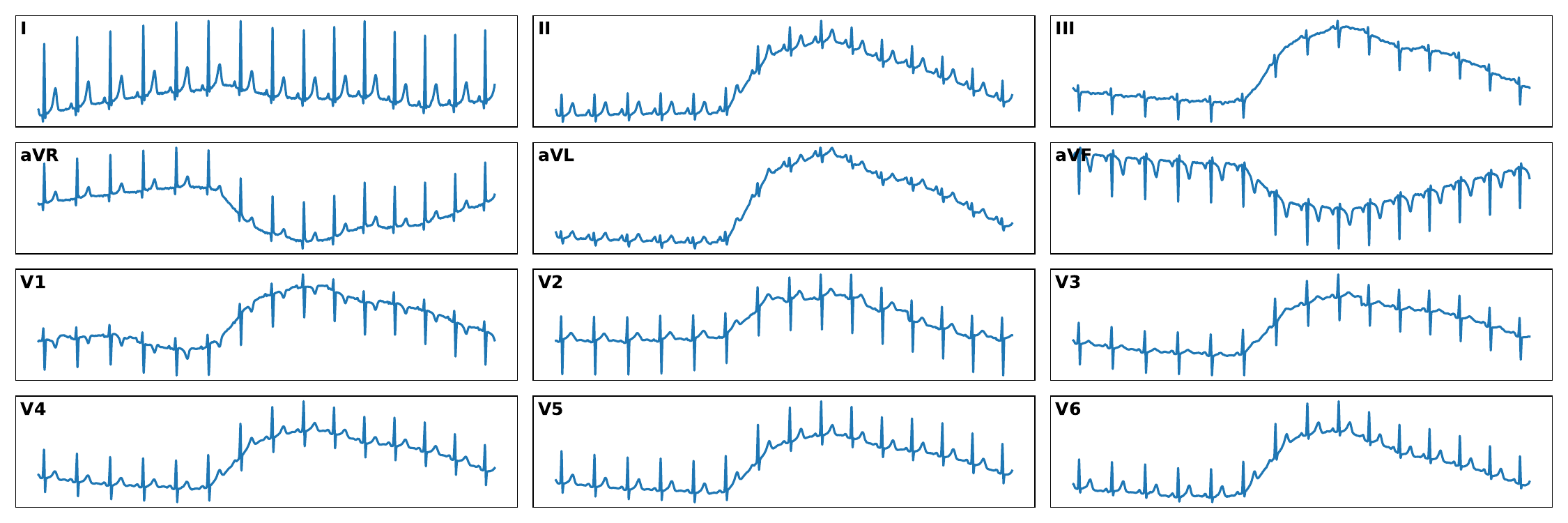}
    \caption{\textbf{Example 12-lead ECG}. Sample record from the CODE-15 dataset \citep{ribeiro2020automatic}.}
    \label{12lead}
\end{figure*}

Our main contributions are as follows:

\begin{itemize}
    \item We propose an ECG-specific JEPA framework for multi-lead ECG representation learning with synchronized temporal masking, a lead-wise predictor, and Cross-Pattern Attention (CroPA).

    \item We provide broad empirical evidence that ECG-JEPA learns transferable representations across linear evaluation, fine-tuning, reduced-lead evaluation, low-shot learning, and noisy settings.

    \item Beyond diagnostic classification, we show that the learned
representations can predict ECG features and support ECG segmentation.
\end{itemize}

In summary, ECG-JEPA introduces a robust SSL framework for 12-lead ECG analysis, overcoming traditional SSL limitations with clinically inspired design elements, scalable architecture, and demonstrated effectiveness on a wide range of tasks. 
Our code is available at \url{https://github.com/sehunfromdaegu/ECG_JEPA}.

\section{Background} 
Self-Supervised Learning (SSL) facilitates learning abstract representations from input data without the need for labeled data, which is particularly beneficial in medical domains where labeled data is scarce and costly to obtain. SSL leverages inherent data patterns to learn useful representations, allowing models to adapt to various downstream tasks with greater robustness to data imbalances \citep{ssl_robust}. We begin in Section \ref{sec:ecg} with an overview of the ECG and its key features, highlighting the critical characteristics essential for understanding ECG data. In Sections \ref{sec:architectures} and \ref{sec:related_works}, we briefly explain key SSL techniques and their specific applications to ECG, respectively.

\subsection{Electrocardiogram (ECG)} \label{sec:ecg}
Electrocardiography is a non-invasive diagnostic method that records the heart's electrical activity over time using electrodes placed on the skin. The result of this recording is called an electrocardiogram (ECG), which visually represents the electrical activity of the heart as a waveform. The standard 12-lead ECG captures electrical activity of the heart from multiple angles. These 12 leads are categorized into limb leads (I, II, III), augmented limb leads (aVR, aVL, aVF), and chest leads (V1-V6). Each lead provides unique information about the heart's electrical activity, offering a comprehensive view that aids in diagnosing various cardiac conditions. Refer to Figure \ref{12lead} for an illustration of 12-lead ECG. 

\begin{figure}[h]
\centering
\includegraphics[scale=0.7]{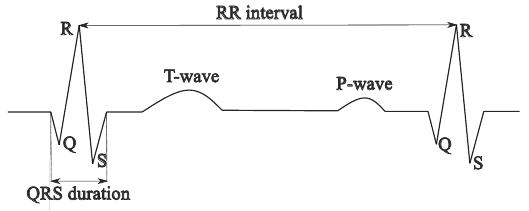}
\caption{\textbf{Key ECG features}. Illustration of P wave, QRS complex, T wave, and interval measurements used in this work.}
\label{ecg_features}
\end{figure}

Key ECG features include heart rate, QRS duration, PR interval, QT interval, and ST segment. These features are identified by measuring specific time intervals or amplitude levels in the ECG waveform. For instance, heart rate is calculated using the formula \(1000 \times (60/\text{RR interval})\) in beats per minute (bpm), where the RR interval is measured in milliseconds (ms). Refer to Figure \ref{ecg_features} for a visual representation of these features.

In this work, we use only 8 leads (I, II, V1-V6) as the remaining 4 leads (III, aVR, aVL, aVF) can be derived from linear combinations of the 8 leads following the \textit{Einthoven's law} \citep{thaler2021only}:

\[
\text{III} = \text{II} - \text{I}, \qquad 
\text{aVR} = - (\text{I} + \text{II})/2, \qquad 
\text{aVL} = (\text{I} - \text{II})/2, \qquad
\text{aVF} = (\text{II} - \text{I})/2.
\]
This choice maintains the necessary diagnostic information while optimizing computational efficiency. A performance comparison between the 8-lead model and the 12-lead model is provided in \ref{sec:12leadcomparison}, demonstrating that the 8-lead model achieves comparable results with reduced computational requirements.

\subsection{Self-Supervised Learning Architectures}\label{sec:architectures}
Self-supervised learning can be broadly categorized into contrastive and non-contrastive methods. Non-contrastive methods can be further divided into generative and non-generative architectures. See \citep{cookbook} for a broader introduction to SSL.

In \textit{contrastive learning}, the model is encouraged to produce similar representations for semantically related inputs \(x^{'}\) and \(x^{''}\), while pushing apart the representations of unrelated inputs \(x^{'}\) and \(y^{'}\). \textit{SimCLR} \citep{simclr} is one of the most popular contrastive methods, using two different augmentations of a single input \(x\) to form semantically similar pairs \(x^{'}\) and \(x^{''}\).

Beyond contrastive methods, \textit{generative architectures} have been particularly successful in recent large language models \citep{bert, gpt3, llama} and in computer vision \citep{mae}. Generative architectures typically involve reconstructing a sample \(x\) from its degraded version \(x'\), employing either encoder-decoder frameworks or other paradigms like decoder-only or encoder-only models. The premise is that reconstructing clean data from a corrupted version reflects the model's deep understanding of the underlying data structure. In encoder-decoder frameworks, the encoder maps the perturbed input \(x'\) into a latent representation, which the decoder then uses to reconstruct the original input \(x\) \citep{dae}. Recently, \citep{balestriero2024} observed that generative architectures prioritize learning principal subspaces of the data, which may limit their capacity to capture semantic representations for perceptual tasks.

As an alternative, \textit{non-generative methods} have shown promise across domains, including computer vision \citep{byol, vicreg, simsiam, ijepa} and video analysis \citep{vjepa}. Among these, the Joint-Embedding Predictive Architecture (JEPA) \citep{path} processes  an input pair \(x\) and its corrupted versions \(x'\) to obtain representations \(z\) and \(z'\) through encoders. Unlike generative architectures that make predictions in the input space, JEPA performs prediction in the latent space by reconstructing \(z\) from \(z'\). This approach effectively avoids the challenge of predicting unpredictable details, a common issue in biological signals.

\subsection{Related Works}\label{sec:related_works}
Early ECG SSL studies adapted contrastive objectives such as CPC and SimCLR to multi-lead ECG, improving label efficiency under limited annotation \citep{cpc, simclr, cpcecg}. \citet{clocs} further introduced ECG-specific positive-pair construction across space, time, and patients in the CLOCS framework, termed Contrastive Multi-Segment Coding (CMSC). 

Later work emphasized masking-based pretraining. Inspired by wav2vec 2.0 \citep{wav2vec2}, ECG adaptations introduced local/global pretraining and random lead masking (RLM) \citep{oh2022lead}. In parallel, masked autoencoder (MAE) ECG methods reconstructed masked temporal or lead patches \citep{maefe, yang2022masked, wang2023unsupervised}, including ECG-MAE \citep{hu2023spatiotemporal} and ST-MEM \citep{guiding}. Recent large-scale models combine multiple objectives, including ECG-FM (combining wav2vec, CMSC, and RLM) and KED (knowledge-enhanced signal-language modeling) \citep{mckeen2025ecg, tian2024foundation}. While many prior ECG SSL studies mainly report diagnostic classification, to our knowledge this is the first work to jointly evaluate ECG feature prediction and ECG segmentation (Sections~\ref{exp:feature} and~\ref{exp:ecg_segmentation}), broadening evaluation beyond diagnosis labels.

\section{Methodology}
\begin{figure*}[t]
    \centering
    \includegraphics[width=\textwidth]{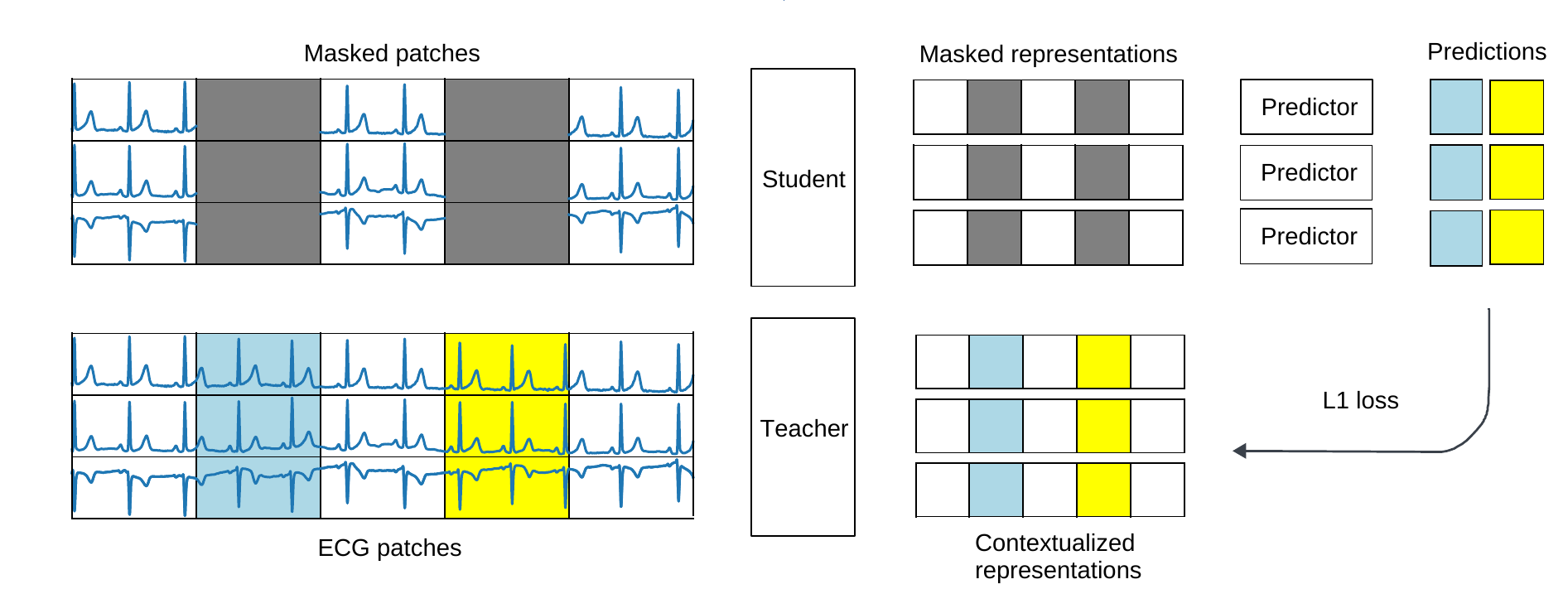}
    \caption{\textbf{ECG-JEPA training overview}. Illustration with \(C = 3\) channels and \(N = 5\) subintervals, where \(I_{vis} = \{1,3,5\}\), representing visible intervals and and \(I_{msk} = \{2,4\}\), representing masked intervals.}
    \label{architecture}
\end{figure*}
ECG-JEPA is trained by predicting masked patches of ECGs in the hidden representation space, using a partial view of the input to infer the missing parts. The proposed architecture utilizes a student-teacher framework, as illustrated in Figure \ref{architecture}. We subdivide the multi-channel ECG into non-overlapping patches and sample a subset of these patches for masking. 

While our model is trained to predict in the representation space, learning by reconstructing the raw signals can be particularly challenging in the ECG domain due to the prevalence of noise. Instead, our model predicts the masked patches in the hidden representation space, where this challenge can be effectively addressed.

Our approach inherently accounts for the presence of noise in biological signals, as the model is trained on raw ECG signals without any preprocessing or noise removal techniques. This design choice ensures that the model is trained on real-world noisy ECG samples (see Figure \ref{12lead}), enabling it to process such signals effectively, even when noise from sources like patient movement or electrical interference is present.

\subsection{Patch Masking}
Let \( x \in \mathbb{R}^{C \times T} \) represent a multi-lead ECG of length \( T \) with \( C \) channels. We divide the interval \([0, T)\) into \( N \) non-overlapping subintervals  of length \( t \). Each subinterval in each channel constitutes a patch \(x_{c,i} \in \mathbb{R}^{t} \) of \( x \), resulting in \(C \times N \) patches \( \{ x_{c,i} \}_{c \in [C], i \in [N]} \), where \([N]\) is the set of integers \(\{1,2,\dots,N\}\). 

The masking strategy in multi-lead ECG must be carefully chosen because patches in different leads at the same temporal position are highly correlated \citep{thaler2021only}, potentially making the prediction task too easy. To address this, we mask all patches across different leads in the same temporal space. With this in mind, we employ two masking strategies: \textit{random masking} and \textit{multi-block masking}.

In random masking, we randomly select a percentage of subintervals to mask, while in multi-block masking, we select multiple consecutive subintervals to mask. Note that we allow these consecutive subintervals to overlap, which requires the model to predict much longer sequences of representations. To evaluate the effectiveness of ECG-JEPA, we use both strategies, with a random masking ratio of \((0.6, 0.7)\) and a multi-block masking ratio of \((0.175, 0.225)\) at a frequency of 4 (see \ref{sec:maskingratio} for an ablation study on varying masking ratios). For either masking strategy, the masking indices are denoted as \(I_{msk} \subset [N]\), and the visible indices as \(I_{vis}\), such that \([N] = I_{msk} \cup I_{vis}\). The unmasked patches \( \{ x_{c,i} \}_{c \in [C], i \in I_{vis}} \) serve as contextual input for the student networks, while the masked patches \( \{ x_{c,i} \}_{c \in [C], i \in I_{msk}} \) are the targets to predict in the representation space.

The patches \( \{ x_{c,i} \}_{c \in [C], i \in [N]} \) are converted into a sequence of token vectors \( \{ x_{c,i}^{\text{tkn}} \}_{c \in [C], i \in [N]} \) of dimension \(D\) using a linear layer, and augmented with positional embeddings. For simplicity, we continue to refer to the token vectors as \(x_{c,i} \in \mathbb{R}^{D}\) with a slight abuse of notation. We employ the conventional 2-dimensional sinusoidal positional embeddings for the student and teacher networks, while 1-dimensional sinusoidal positional embeddings are used for the predictor network.

\subsection{Teacher, Student, and Predictor} \label{sec:teacher,student,predictor}

ECG-JEPA is built upon three key components: the teacher network, the student network, and the predictor network, each playing a distinct role in the model's learning process. The teacher and student networks are based on standard transformer architectures, while the predictor network, a smaller transformer, operates on single-channel representations. Despite operating on single channels, the predictor effectively encodes information from all leads, leveraging the self-attention mechanism to integrate contextual dependencies.

The teacher network handles the entire \(C \times N\) patches \( \{x_{c,i}\}_{c \in [C], i \in [N]} \), generating fully contextualized representations \( \{z_{c,i}\}_{c \in [C], i \in [N]} \). The student network, however, processes only \(C \times Q\) visible (unmasked) patches \( \{x_{c,i}\}_{c \in [C], i \in I_{vis}} \), where \(Q = |I_{vis}|\) represents the number of visible time intervals. 
The representations \( \{z^{\text{std}}_{c,i}\}_{c \in [C], i \in I_{vis} } \) from the student are then concatenated with \(C \times (N-Q\)) (learnable) mask tokens \(z_{msk} \in \mathbb{R}^{D}\), resulting in \(C \times N\) representations. Subsequently, each lead's representations 
\(
\{ z^{\text{std}}_{c,i} \}_{i \in I_{vis}} \cup \{z_{msk}, \dots, z_{msk}\}
\) are passed to the predictor, generating the predictions 
\(
\{\widehat{z_{c,i}}\}_{i \in [N]}
\). 

Finally, the objective function of ECG-JEPA is defined as the L1 distance between the predicted representations for the masked patches and their corresponding teacher-generated representations. Formally,
\[
\mathcal{L} = \sum_{c \in [C]} \frac{1}{|I_{msk}|}\sum_{i \in I_{msk}} \| \widehat{z_{c,i}} - z_{c,i} \|_{1}
\]

The main challenge in the student-teacher framework—or, more generally, in any joint-embedding architecture—is \textit{model collapse}, where both encoders produce constant outputs regardless of their inputs, thereby minimizing the loss function. A common approach to prevent collapse is to update the teacher network's weights using an exponential moving average (EMA) of the student network's weights, which we adopt in our model. The details of EMA are provided in \ref{sec:EMA}.

\begin{figure*}
    \centering
    \includegraphics[scale=0.5]{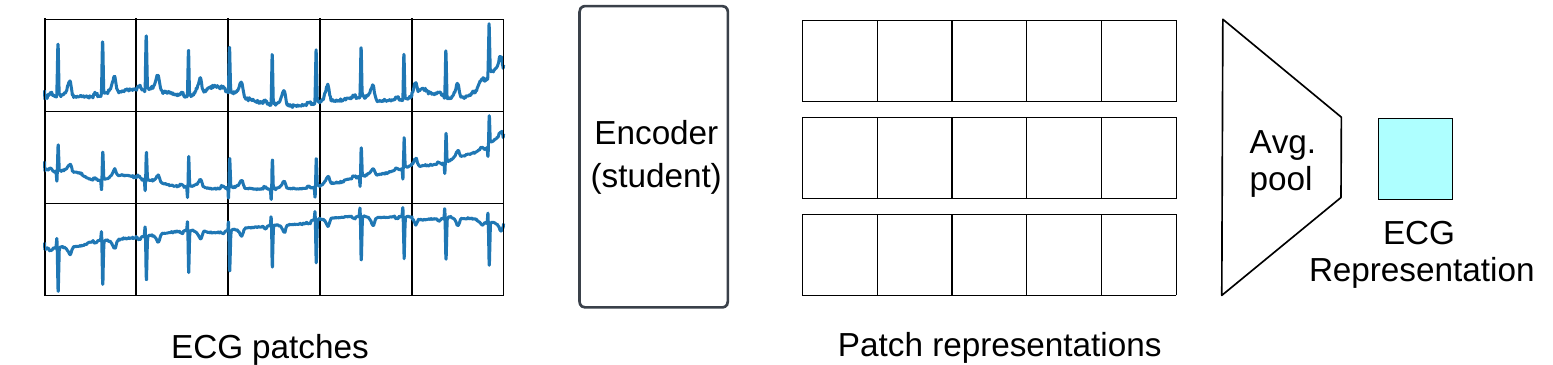}
    \caption{\textbf{Inference procedure}. Patch-level averaging produces a single ECG representation vector (cyan).}
    \label{inference}
\end{figure*}

At inference time, only the student network is used as the encoder. The encoder's outputs are average-pooled to produce the final ECG representation, which serves as the feature vector for downstream tasks. The dimension of this latent representation vector matches the encoder's token dimension, which is set to \(D = 768\) in our case. See Figure \ref{inference} for an illustration.

\subsection{Cross-Pattern Attention (CroPA)}

\begin{figure}[t]
    \centering
    \includegraphics[width=\textwidth]{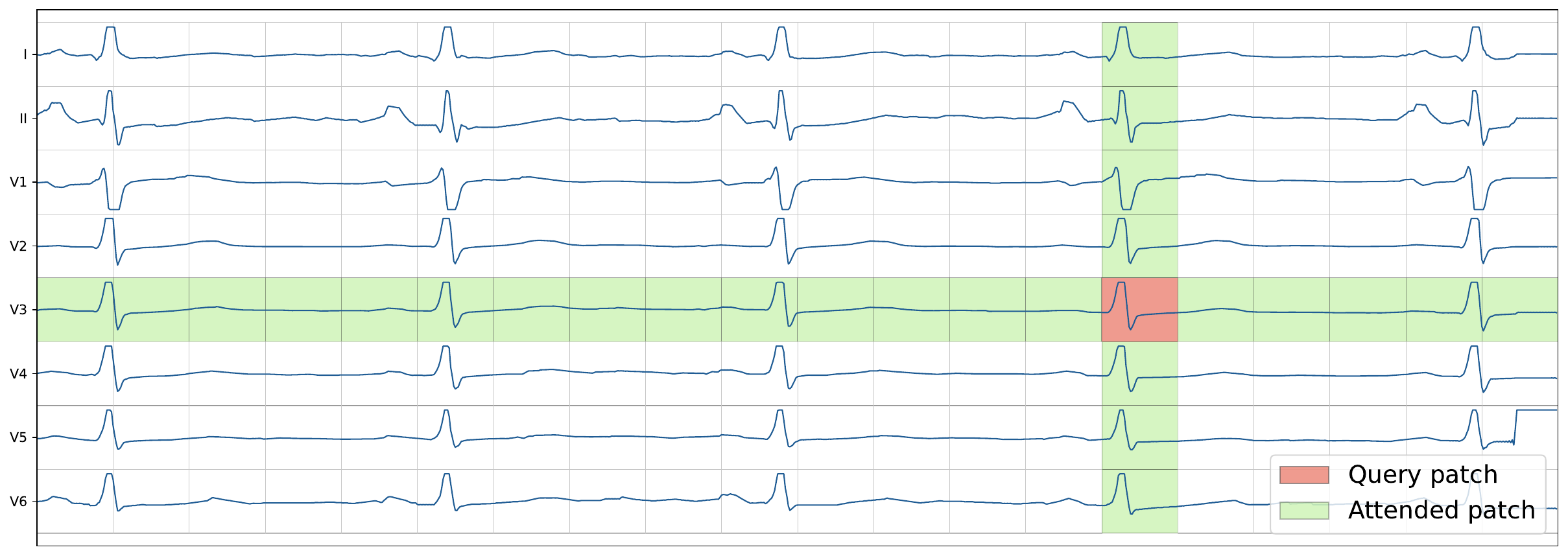}
    \caption{\textbf{CroPA visualization}. Query patch (red) and attended patches (green) illustrate cross-lead contextual attention.}
    \label{cross}
\end{figure}

Interpreting a 12-lead ECG involves analyzing individual leads as well as comparing signals across multiple leads to enhance diagnostic accuracy. Multi-lead comparison not only helps distinguish artifacts from true abnormalities but also provides a more comprehensive assessment of cardiac activity. In \citep{thaler2021only}, for instance, the following quotes illustrate this principle:

\begin{quote}
\textit{`` \dots The diagnosis (of posterior infarction) must therefore be made by looking for reciprocal changes in the anterior leads, for example, a tall R wave in leads V1, V2, or V3.''}
\end{quote}

\begin{quote}
\textit{``\dots ST-segment depression of at least 1 mm in leads V1-V3 if deep S waves are present is strongly suggestive of an evolving infarction.''}
\end{quote}

Motivated by these observations, we introduce Cross-Pattern Attention (CroPA), a masked self-attention mechanism that imposes an inductive bias by restricting attention to clinically relevant patches. Specifically, a patch \(x_{c,i}\) attends to another patch \(x_{c',i'}\) if and only if either (1) they belong to the same lead (\(c = c'\)), or (2) they are in the same temporal space (\(i = i'\)) (see Section \ref{sec:teacher,student,predictor} for notations).

This design aligns with how ECG signals are clinically interpreted, where intra-lead and temporally adjacent signals are most relevant. By incorporating this inductive bias, CroPA focuses on relevant intra-lead relationships, reducing interference from unrelated signals across other channels and temporal spaces. Unlike standard self-attention, which treats all patches equally, CroPA adopts a structured approach that mirrors the clinical interpretation process, leading to improved performance on downstream tasks as demonstrated in Section \ref{exp:cropa}.

\section{Experimental Settings}
For ECG-JEPA experiments, 10-second multi-lead ECG signals were resampled to 250\(\,\mathrm{Hz}\), yielding $T = 2500$ time points. We divided the interval $[0, T)$ into $N = 50$ non-overlapping subintervals, each of length $t = 50$. The model was trained for 100 epochs without data augmentation or noise removal preprocessing, and the final checkpoint was used for downstream tasks. For other baseline models, we followed each original paper's preprocessing and input-setting protocol. Additional experimental details are provided in \ref{sec:Experimental Details}.

\subsection{Pretraining Datasets}
Training SSL models with large datasets is crucial for developing generalized representations. However, most previous works have used relatively small datasets, with the exception of \citep{guiding}, where an SSL model was trained with a large number of 12-lead ECGs. Following \citep{guiding}, we use the \textit{Chapman} \citep{chapman}, \textit{Ningbo} \citep{ningbo}, and \textit{CODE-15} \citep{chen2019large} datasets for pretraining ECG-JEPA. The Chapman and Ningbo datasets collectively consist of 45,152 10-second 12-lead ECGs at 500\(\,\mathrm{Hz}\). CODE-15 includes 345,779 12-lead ECGs from 233,770 patients at 400\(\,\mathrm{Hz}\), with 143,328 being 10-second recordings. After excluding recordings with missing values, we have 43,240 ECGs from Chapman and Ningbo and 130,900 ECGs from CODE-15.

\subsection{Downstream Datasets}
To evaluate the performance of ECG-JEPA on downstream tasks, we use \textit{PTB-XL}~\citep{ptbxl}, \textit{CPSC2018}~\citep{cpsc}, and \textit{G12EC}~\citep{cinc2020}. See \ref{sec:dataset_details} for detailed split statistics.

\textbf{\textit{PTB-XL}} contains 21,837 12-lead ECG recordings sampled at 500 Hz. It provides three label categories: \textit{diagnostic}, \textit{rhythm}, and \textit{form}. In our main experiments, we use the diagnostic labels grouped into five superclasses. Unless otherwise noted, all references to \textit{PTB-XL} in this paper refer to the diagnostic label set. For visualization, we also use rhythm labels in ~\ref{sec:visualization}. We follow the official fold-based split protocol.

\textbf{\textit{CPSC2018}} comprises 6,877 12-lead ECG recordings annotated with nine cardiac conditions. The dataset is provided in seven folds; we use the first five folds for training and the remaining two folds for validation and test. This dataset contains 9 classes.

\textbf{\textit{G12EC}} (Georgia 12-lead ECG Challenge dataset) contains 10,292 12-lead ECG recordings sampled at 500 Hz. We follow the PhysioNet Challenge 2020 scoring protocol~\citep{cinc2020} and exclude classes with fewer than 50 samples in the full dataset (train, validation, and test combined). This yields 21 labels for multi-label tasks and 15 classes for multi-class tasks; detailed processing steps are provided in \ref{sec:dataset_details}.

\subsection{Architecture}
Our model employs transformer encoder architectures for the student, teacher, and predictor networks. Both the teacher and student networks consist of 12 layers with 16 attention heads and a hidden dimension of 768. The predictor network, designed as a smaller transformer encoder, comprises 6 layers with 12 attention heads and a hidden dimension of 384. While the teacher and student networks process the multi-lead ECG data holistically, the predictor operates on each lead independently to reconstruct the masked representations. Importantly, this does not imply that the predictor relies solely on single-lead information for the reconstruction task; due to the self-attention mechanism, the input representations for each lead still encapsulate information from all leads.

\section{Experiments}\label{sec:experiments}
In this section, we evaluate the performance of the learned representations across various downstream tasks to demonstrate their generalizability and ability to capture essential ECG features. ECG-JEPA is compared against several state-of-the-art self-supervised learning (SSL) methods.

For classification tasks, we use AUC (Area Under the ROC Curve) and F1 scores as evaluation metrics. AUC provides a comprehensive measure of discriminative ability by considering performance across all classification thresholds, making it more robust to variations in decision boundaries. In contrast, the F1 score balances precision and recall at a fixed threshold, offering insights into the model’s performance when a specific decision boundary is chosen. We select the best model configuration based on AUC performance on the validation set, and all reported results are computed on the held-out test set.

In multi-label classification, AUC is computed by averaging the scores from binary classification for each label, whereas in multi-class classification, AUC is calculated using the one-vs-rest approach. For both tasks, F1 scores are macro-averaged across all classes to ensure equal weighting. For brevity, ECG-JEPA\(_{rb}\) and ECG-JEPA\(_{mb}\) denote ECG-JEPA models trained using random masking and multi-block masking strategies, respectively.

For linear evaluations, we performed a grid search over 10 logarithmically spaced learning rates between \(10^{-1}\) and \(10^{-4}\). For fine-tuning, we used 10 logarithmically spaced base learning rates between \(10^{-2}\) and \(10^{-5}\), where the actual learning rate is computed as \( \textit{lr} = \textit{base\_lr} \times \textit{batchsize} / 256 \), following the heuristic proposed by \citep{baselr}. The best-performing learning rate was selected based on validation set performance, and all reported scores reflect evaluation on the held-out test set.

We first present classification results, including linear evaluation, reduced-lead evaluation, low-shot learning, and fine-tuning. We then examine whether the learned representations preserve clinically meaningful signal structure through ECG feature extraction, segmentation, and robustness under noise, and finally assess the effect of CroPA.

 \begin{table*}[t]
    \centering
    \caption{\textbf{Linear evaluation across datasets}. AUC is reported for multi-label and multi-class tasks on \textit{PTB-XL}, \textit{CPSC2018}, and \textit{G12EC}.}
    \label{tab:linear eval}
    \footnotesize
    \setlength{\tabcolsep}{4pt}
    \renewcommand{\arraystretch}{1.08}
    \begin{adjustbox}{max width=\textwidth}
    \begin{threeparttable}
    \begin{tabular}{lcccccc}
    \toprule
    \multirow{2}{*}{Method} & \multicolumn{3}{c}{\textbf{Multi-label AUC}} & \multicolumn{3}{c}{\textbf{Multi-class AUC}} \\
    \cmidrule(lr){2-4}\cmidrule(lr){5-7}
    & \textit{PTB-XL} & \textit{CPSC2018} & \textit{G12EC} & \textit{PTB-XL} & \textit{CPSC2018} & \textit{G12EC} \\
    \midrule
    MoCo v3\tnote{1}  & -     & -     & -                        & 0.739 & 0.712 & -              \\
    MTAE\tnote{1}     & -     & -     & -                        & 0.807 & 0.818 & -              \\
    MLAE\tnote{1}     & -     & -     & -                        & 0.779 & 0.794 & -              \\
    ST-MEM            & 0.882 & 0.955 & \underline{0.893}           & 0.879 & 0.964 & \textbf{0.910} \\
    SimCLR            & 0.875 & 0.915 & 0.859                    & 0.830 & 0.925 & 0.862              \\
    ECG-FM            & 0.878 & 0.916 & 0.865                    & 0.856 & 0.931 & 0.869          \\
    KED         & 0.885 & 0.883 & 0.819                    & 0.888 & 0.906 & 0.768          \\
    \midrule
    ECG-JEPA\(_{rb}\) & \underline{0.906} & \underline{0.965} & 0.882 & \underline{0.897} & \underline{0.970} & 0.899 \\
    ECG-JEPA\(_{mb}\) & \textbf{0.912}    & \textbf{0.966}    & \textbf{0.895} & \textbf{0.903}    & \textbf{0.973}    &
\underline{0.908} \\
    \bottomrule
    \end{tabular}
    \begin{tablenotes}
    \footnotesize
    \item[1] Scores reported in \citep{guiding}; unavailable entries are marked as ``-''.
    \end{tablenotes}
    \end{threeparttable}
    \end{adjustbox}
\end{table*}

\subsection{Linear Evaluation}\label{sec:linear eval}
Table \ref{tab:linear eval} presents the results of our linear evaluation on the \textit{PTB-XL}, \textit{CPSC2018}, and \textit{G12EC} datasets. We train a linear classifier on top of the frozen representations for 10 epochs and evaluate its performance on downstream tasks. Further training beyond 10 epochs does not lead to any significant improvement in performance. As shown in the table, ECG-JEPA consistently outperforms other SSL methods, demonstrating superior efficiency and effectiveness with substantially reduced computational resources.

\subsection{Reduced Lead Evaluation}\label{sec:reduced lead}
To evaluate ECG-JEPA's performance under reduced input settings, we leveraged the flexibility of transformer architectures to handle variable input lengths. In this experiment, we conducted a linear evaluation on the \textit{PTB-XL} multi-label task using only a single lead (Lead II) and two leads (Lead II and V1), training linear classifiers as in Section \ref{sec:linear eval}. Table \ref{tab:reduced_lead} presents the results. Notably, ECG-JEPA maintains strong performance even with fewer leads, which is valuable for practical applications in mobile health monitoring.

\begin{table}[t]
    \centering
    \caption{\textbf{Reduced lead evaluation}. Linear evaluation of \textit{PTB-XL} multi-label classification with single-lead (II) and dual-lead (II and V1) inputs.}
    \label{tab:reduced_lead}
    \footnotesize
    \setlength{\tabcolsep}{4pt}
    \renewcommand{\arraystretch}{1.08}
    \begin{tabular}{lcccc}
        \toprule
        \multirow{2}{*}{Method} & \multicolumn{2}{c}{1-Lead} & \multicolumn{2}{c}{2-Lead} \\
        \cmidrule(lr){2-3}\cmidrule(lr){4-5}
        & AUC & F1 & AUC & F1 \\
        \midrule
        ST-MEM            & 0.831             & 0.540             & 0.857             & 0.564 \\
        ECG-FM            & 0.837             & \underline{0.560} & 0.850             & 0.572 \\
        KED         & 0.803             & 0.437             & 0.835             & 0.492 \\
        ECG-JEPA\(_{rb}\) & \underline{0.845} & 0.558             & \underline{0.878} & \underline{0.606} \\
        ECG-JEPA\(_{mb}\) & \textbf{0.849}    & \textbf{0.570}    & \textbf{0.879}    & \textbf{0.641} \\
        \bottomrule
    \end{tabular}
\end{table}

\subsection{Low-shot Linear Evaluation}\label{sec:lowshot}
Table \ref{tab3:multilabel lowshot} presents the performance comparison on the low-shot task. Low-shot learning is particularly challenging, as models must generalize effectively with limited labeled data. Given the difficulty and resource-intensive nature of obtaining labeled data in medical research, low-shot learning represents a realistic and critical scenario in the medical field. In this experiment, we evaluate the performance of ECG-SSL models on the \textit{PTB-XL} multi-label task using only 1\% and 10\% of the training and validation sets, while keeping the entire test set fixed. As shown in the table, ECG-JEPA demonstrates a clear advantage over other SSL methods, with its effectiveness becoming particularly evident in low-shot learning tasks. This suggests that ECG-JEPA can be particularly well-suited for transfer learning where labeled data is scarce.

\begin{table}[t]
    \centering
    \caption{\textbf{Low-shot linear evaluation}. Macro AUC mean and standard deviation on \textit{PTB-XL} multi-label classification using 1\% and 10\% training data (three independent samplings).}
    \label{tab3:multilabel lowshot}
    \footnotesize
    \setlength{\tabcolsep}{4pt}
    \renewcommand{\arraystretch}{1.08}
    \begin{tabular}{lcc}
        \toprule
        Method & 1\% & 10\% \\
        \midrule
        ST-MEM            & 0.817 \(\pm\) 0.000 & 0.865 \(\pm\) 0.001 \\
        SimCLR            & 0.792 \(\pm\) 0.006 & 0.852 \(\pm\) 0.002 \\
        ECG-FM            & 0.729 \(\pm\) 0.013 & 0.844 \(\pm\) 0.001 \\
        KED               & 0.772 \(\pm\) 0.010 & 0.841 \(\pm\) 0.002 \\
        \midrule
        ECG-JEPA\(_{rb}\) & \textbf{0.839 \(\pm\) 0.002} & \underline{0.887 \(\pm\) 0.001} \\
        ECG-JEPA\(_{mb}\) & \underline{0.836 \(\pm\) 0.003} & \textbf{0.893 \(\pm\) 0.001} \\
        \bottomrule
    \end{tabular}
\end{table}

\subsection{Fine-tuning}\label{sec:fine_tuning}
Fine-tuning evaluates the quality of learned representations by testing the model's ability to adapt pre-trained features to downstream tasks. We append a linear classification head to the encoder and train the entire network for up to 100 epochs, using early stopping with a patience of 10 based on the validation set. As previously mentioned, we evaluated 10 logarithmically spaced base learning rates between \(10^{-2}\) and \(10^{-5}\), and selected the best model using validation AUC.

To further boost performance during fine-tuning, preprocessing steps are applied to both training and test sets. These steps include high-pass and low-pass filtering to mitigate common ECG artifacts such as baseline drift and powerline interference.

Table \ref{tab4:multiclass finetune} presents fine-tuning results (AUC) for both multi-label and multi-class tasks on \textit{PTB-XL}, \textit{CPSC2018}, and \textit{G12EC}. ECG-JEPA is compared with other SSL methods and a supervised baseline.

\begin{table*}[t]
    \centering
    \begin{threeparttable}
        \caption{\textbf{Fine-tuning performance across datasets}. AUC on multi-label and multi-class tasks for \textit{PTB-XL}, \textit{CPSC2018}, and \textit{G12EC}.}
        \label{tab4:multiclass finetune}
        \footnotesize
        \setlength{\tabcolsep}{4pt}
        \renewcommand{\arraystretch}{1.08}
        \begin{tabular}{lcccccc}
            \toprule
            \multirow{2}{*}{Method} & \multicolumn{3}{c}{\textbf{Multi-label AUC}} & \multicolumn{3}{c}{\textbf{Multi-class AUC}} \\
            \cmidrule(lr){2-4}\cmidrule(lr){5-7}
            & \textit{PTB-XL} & \textit{CPSC2018} & \textit{G12EC} & \textit{PTB-XL} & \textit{CPSC2018} & \textit{G12EC} \\
            \midrule
            Supervised            & 0.878 & 0.884 & 0.796 & 0.882 & 0.892 & 0.804 \\
            MoCo v3\tnote{1}      & - & - & - & 0.913 & 0.967 & - \\
            MTAE\tnote{1}         & - & - & - & 0.910 & 0.961 & - \\
            MLAE\tnote{1}         & - & - & - & 0.915 & 0.973 & - \\
            ST-MEM                & \underline{0.929} & \textbf{0.973} & \textbf{0.915} & 0.910 & \underline{0.977} & \textbf{0.949} \\
            SimCLR                & 0.918 & 0.936 & 0.865 & \underline{0.928} & 0.955 & 0.861 \\
            ECG-FM                & 0.899 & 0.922 & 0.825 & 0.895 & 0.947 & 0.796 \\
            KED             & 0.901 & 0.891 & 0.809 & 0.906 & 0.923 & 0.848 \\
            \midrule
            ECG-JEPA\(_{rb}\)     & \textbf{0.931} & \textbf{0.973} & \underline{0.906} & \underline{0.928} & 0.976 & \underline{0.940} \\
            ECG-JEPA\(_{mb}\)     & 0.926 & 0.969 & 0.884 & \textbf{0.934} & \textbf{0.980} & 0.938 \\
            \bottomrule
        \end{tabular}
        \begin{tablenotes}
            \footnotesize
            \item[1] Scores reported in \citep{guiding}; unavailable entries are marked as ``-''.
        \end{tablenotes}
    \end{threeparttable}
\end{table*}

\subsection{ECG Feature Extraction} \label{exp:feature}
\begin{table}[t]
    \centering
    \caption{\textbf{ECG feature regression}. Mean absolute errors on the \textit{PTB-XL} normal-sample test split; test-set mean heart rate and QRS duration are 69.67 BPM (\(\pm 12.92\)) and 90.34 ms (\(\pm 6.23\)).}
    \label{tab5:performance_comparison}
    \footnotesize
    \setlength{\tabcolsep}{4pt}
    \renewcommand{\arraystretch}{1.08}
    \begin{tabular}{lcc}
        \toprule
        \multicolumn{1}{c}{} & \multicolumn{2}{c}{Mean Absolute Error} \\
        \cmidrule(lr){2-3}
        Method & Heart Rate (BPM) & QRS Dur. (ms) \\
        \midrule
        ST-MEM            & 0.68 \small{\(\pm\) 0.68} & \textbf{1.42 \small{\(\pm\) 1.17}} \\
        SimCLR            & 1.62 \small{\(\pm\) 2.13} & 2.13 \small{\(\pm\) 2.64} \\
        ECG-FM            & 2.67 \small{\(\pm\) 2.84} & \underline{1.73 \small{\(\pm\) 1.50}} \\
        KED         & 1.16 \small{\(\pm\) 1.82} & 2.64 \small{\(\pm\) 2.21} \\
        \midrule
        ECG-JEPA\(_{rb}\) & \underline{0.50 \small{\(\pm\) 0.67}} & 1.89 \small{\(\pm\) 1.51} \\
        ECG-JEPA\(_{mb}\) & \textbf{0.40 \small{\(\pm\) 0.67}} & 1.98 \small{\(\pm\) 1.54} \\
        \bottomrule
    \end{tabular}
\end{table}

Beyond classification, we assess whether the frozen pretrained representations preserve key ECG features such as heart rate and average QRS duration by training a linear regression layer on top. Unlike classification tasks, which focus on categorical patterns, these features are directly tied to the signal's morphology.

Various methods exist for segmenting ECG signals \citep{sereda2019ecg, moskalenko2020deep, chen2023post, segmentation}, which can be used to extract ECG features. For this experiment, we used a publicly available segmentation model \citep{segmentation} to generate ground-truth labels for heart rate and QRS duration from the \textit{PTB-XL} dataset.

To compute heart rate, the segmentation model identifies R peaks and calculates the average RR interval across the ECG signal. The heart rate is then derived using the formula \(1000 \times \left(60/\text{avg RR interval}\right)\), where the RR interval is expressed in milliseconds.

For average QRS duration, the segmentation model detects the onset and offset of each QRS interval within the ECG. The duration of each QRS interval is computed as the difference between its offset and onset. The average QRS duration is then calculated as the mean of all detected QRS durations. We then trained a linear regression model on the learned representations to predict these features, using mean squared error (MSE) as the loss function. The best-performing model was selected based on validation MSE, and all reported results are computed on the held-out test set.

Table \ref{tab5:performance_comparison} reports the means and standard deviations of the absolute differences between the predicted and extracted values for heart rate and QRS duration across the \textit{PTB-XL} test set.

\subsection{ECG Segmentation} \label{exp:ecg_segmentation}
\begin{figure}[t]
    \centering
    \begin{subfigure}{\linewidth}
        \centering
        \includegraphics[width=\linewidth]{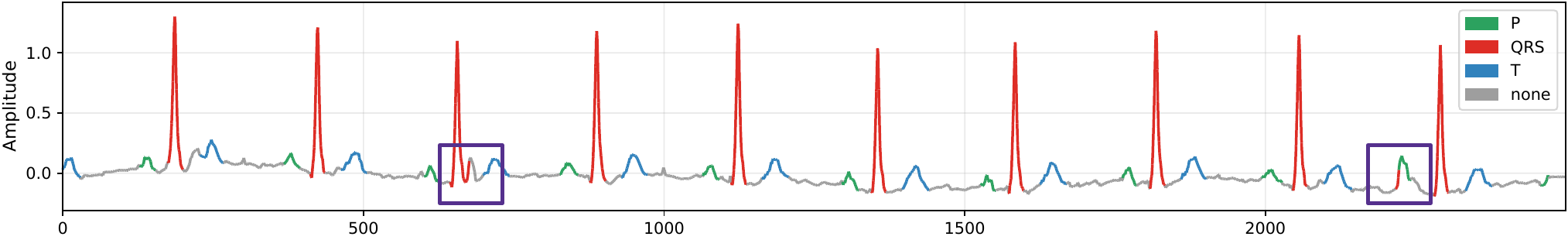}
        \caption{Pseudo label.}
    \end{subfigure}

    \vspace{0.4em}
    \begin{subfigure}{\linewidth}
        \centering
        \includegraphics[width=\linewidth]{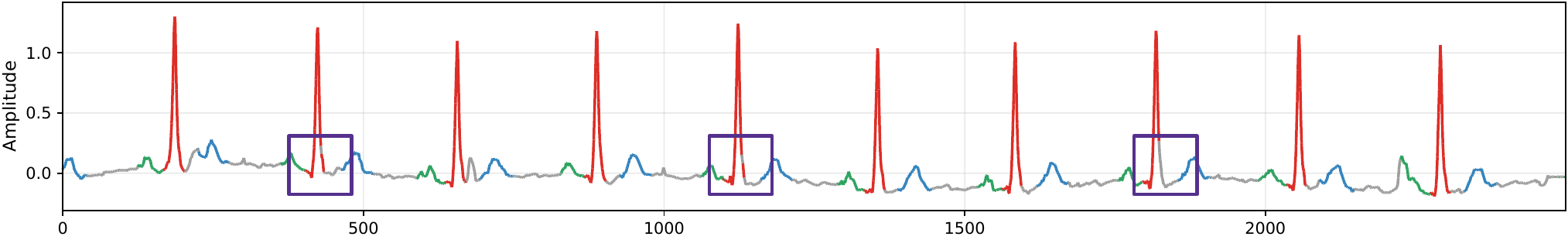}
        \caption{After epoch 1.}
    \end{subfigure}

    \vspace{0.4em}
    \begin{subfigure}{\linewidth}
        \centering
        \includegraphics[width=\linewidth]{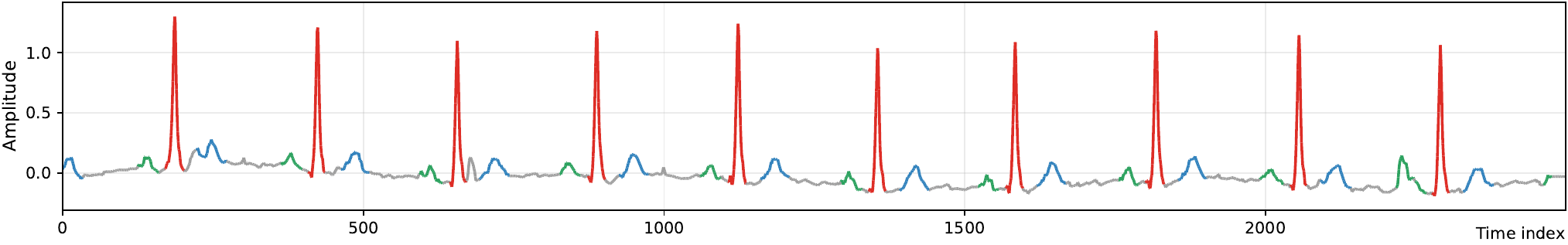}
        \caption{After epoch 10.}
    \end{subfigure}

    \caption{\textbf{Example ECG segmentation with frozen ECG-JEPA$_{rb}$}. (a) Pseudo
    labels; boxed regions indicate labeling errors. (b) Linear-head predictions after
    1 training epoch; boxed regions indicate incorrect predictions, including
    discontinuous QRS complexes and incorrect QRS prediction. (c) Linear-head
    predictions after 10 training epochs, which appear to correct several pseudo-label
    errors.}
    \label{fig:ecg_segmentation_examples}
\end{figure}

We further evaluate whether ECG-JEPA preserves fine-grained ECG morphology using lead-specific segmentation on \textit{PTB-XL}. Segmentation labels are generated with the same publicly available segmentation model \citep{segmentation} used in Section~\ref{exp:feature}. A linear head \(h: \mathbb{R}^{768} \to \mathbb{R}^{50 \times 4}\) is appended to the encoder to predict one of four segmentation classes at each timestamp. We perform both linear evaluation and fine-tuning for this task. For linear evaluation, we train only the linear head for 10 epochs. For fine-tuning, we train the full network for up to 100 epochs with early stopping (patience 10), following the same protocol as Sections~\ref{sec:linear eval} and~\ref{sec:fine_tuning}.

We report class-wise IoU and mean IoU (mIoU), where
\[
\mathrm{IoU}_c = \frac{TP_c}{TP_c + FP_c + FN_c}, \qquad
\mathrm{mIoU} = \frac{1}{4}\sum_{c}\mathrm{IoU}_c .
\]

\begin{table}[t]
    \centering
    \caption{\textbf{ECG segmentation performance}. Lead-specific IoU and mIoU on \textit{PTB-XL} normal signals with four classes (\(P,\ QRS,\ T,\ none\)).}
    \label{tab:ecg_segmentation}
    \footnotesize
    \setlength{\tabcolsep}{4pt}
    \renewcommand{\arraystretch}{1.08}
    \begin{tabular}{llccccc}
        \toprule
        Method & Setting & mIoU & IoU\(_P\) & IoU\(_{QRS}\) & IoU\(_T\) & IoU\(_{none}\) \\
        \midrule
        ECG-JEPA\(_{rb}\) & Linear (frozen) & 0.888 & 0.828 & 0.890 & 0.929 & 0.906 \\
        ECG-JEPA\(_{mb}\) & Linear (frozen) & 0.890 & 0.828 & 0.891 & 0.932 & 0.907 \\
        ECG-JEPA\(_{rb}\) & Fine-tuning & \textbf{0.954} & \textbf{0.924} & \textbf{0.958} & \underline{0.970} & \textbf{0.964} \\
        ECG-JEPA\(_{mb}\) & Fine-tuning & \underline{0.952} & \underline{0.921} & \underline{0.955} & \textbf{0.971} & \underline{0.963} \\
        \bottomrule
    \end{tabular}
\end{table}

Figure~\ref{fig:ecg_segmentation_examples} provides a qualitative example of ECG segmentation with a frozen ECG-JEPA$_{rb}$ encoder, and Table~\ref{tab:ecg_segmentation} shows that fine-tuning substantially improves performance over linear evaluation for both ECG-JEPA variants, with ECG-JEPA\(_{rb}\) achieving the best overall mIoU.

\subsection{Robustness Under Noise}\label{exp:robustness}
We further evaluate robustness under three noisy scenarios: (1) with basic preprocessing applied to remove noise (noise level 0), (2) without preprocessing, retaining the inherent noise present in raw signals (noise level 1), and (3) with artificially introduced noise (noise level 2). Figure~\ref{noise} in \ref{app:noise} visualizes the three settings.

We compare ECG-JEPA\(_{rb}\), ECG-JEPA\(_{mb}\), ECG-FM, KED, SimCLR, and ST-MEM on \textit{PTB-XL} multi-label linear evaluation. To simulate realistic noise, we incorporate two common ECG artifacts, baseline drift and powerline interference, using simple mathematical models. Detailed preprocessing and noise-generation procedures are provided in \ref{app:noise}.

\begin{figure}[t]
    \centering
    \includegraphics[width=0.8\textwidth]{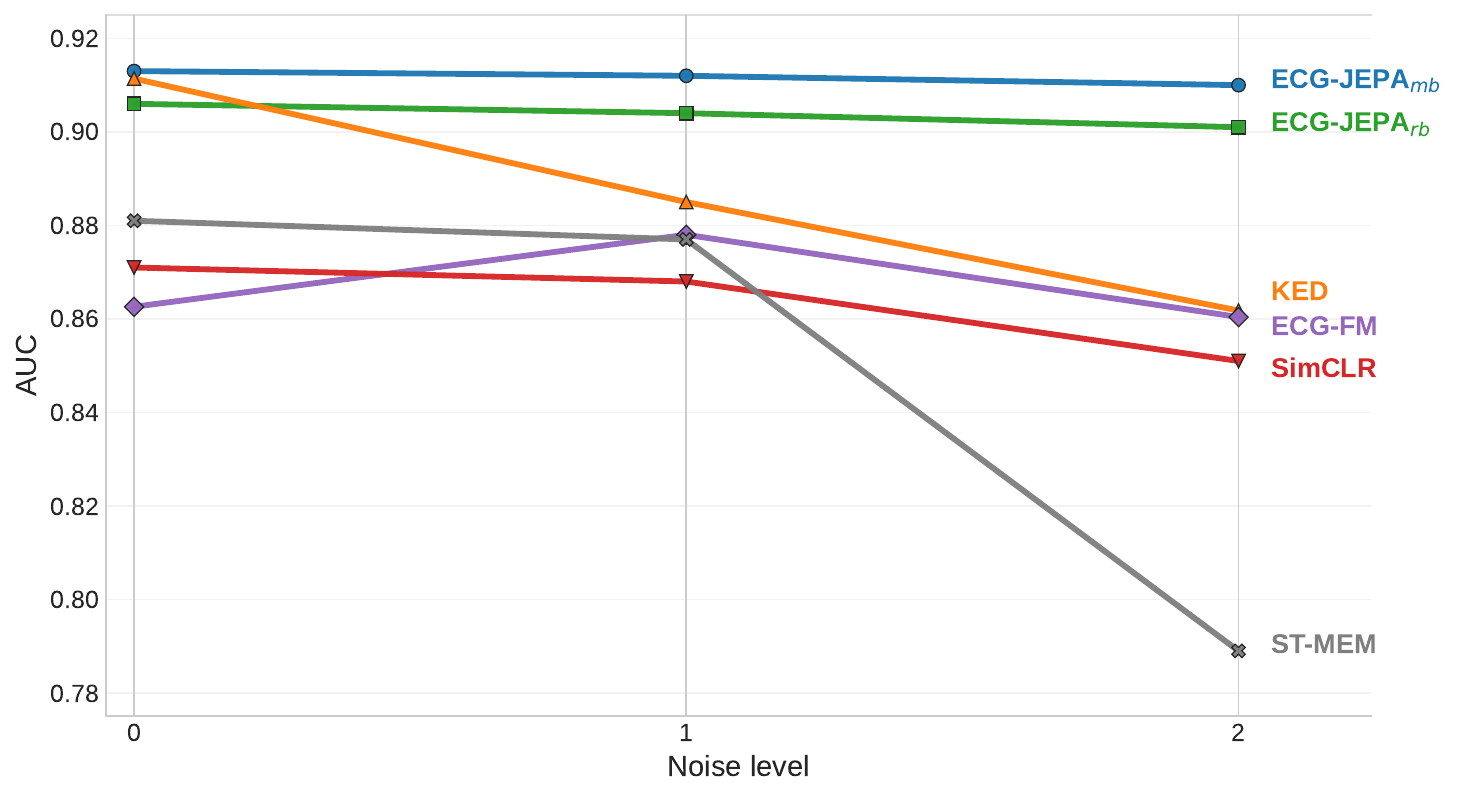}
    \caption{\textbf{Noise robustness}. AUC under noise levels 0, 1, and 2 on \textit{PTB-XL} multi-label linear evaluation.}
    \label{fig:noise_comparison}
\end{figure}

Figure~\ref{fig:noise_comparison} shows that the gap between methods becomes larger as noise increases. At noise level 0, KED slightly outperforms ECG-JEPA\(_{rb}\), but its performance drops more sharply under stronger corruption. In contrast, both ECG-JEPA variants remain comparatively stable across the three settings, with ECG-JEPA\(_{mb}\) achieving the best AUC throughout. Overall, these results indicate that ECG-JEPA learns representations that remain robust under common ECG artifacts.

\subsection{Effect of CroPA}\label{exp:cropa}

Table \ref{tab6:cropa} presents the results of our evaluation of the effectiveness of CroPA. CroPA introduces a clinically motivated inductive bias that mirrors diagnostic practice and enables more efficient learning from multi-lead ECG data. Without CroPA, models may require more epochs to converge to comparable performance.

To assess the contribution of CroPA, we trained ECG-JEPA with and without CroPA for 100 pretraining epochs and evaluated their performance on multi-class classification across \textit{PTB-XL}, \textit{CPSC2018}, and \textit{G12EC}, using both linear evaluation and fine-tuning. CroPA consistently improves performance, indicating that it effectively captures inter-lead dependencies and helps the model learn more meaningful representations. Further statistical analysis on the effect of CroPA is provided in Section \ref{statistical analysis on cropa}.

\begin{table*}[t]
    \centering
    \caption{\textbf{Effect of CroPA}. Multi-class AUC of ECG-JEPA with 100-epoch pretraining across \textit{PTB-XL}, \textit{CPSC2018}, and \textit{G12EC}.}
    \label{tab6:cropa}
    \footnotesize
    \setlength{\tabcolsep}{4pt}
    \renewcommand{\arraystretch}{1.08}
    \begin{tabular}{lccccccc}
        \toprule
        \multirow{2}{*}{Mask} & \multirow{2}{*}{CroPA} &
        \multicolumn{2}{c}{\textit{PTB-XL}} &
        \multicolumn{2}{c}{\textit{CPSC2018}} &
        \multicolumn{2}{c}{\textit{G12EC}} \\
        \cmidrule(lr){3-4}\cmidrule(lr){5-6}\cmidrule(lr){7-8}
        & & lin & ft & lin & ft & lin & ft \\
        \midrule
        Random      & \(\times\)     & 0.888 & 0.923 & \textbf{0.970} & \textbf{0.976} & 0.890 & 0.924 \\
        Random      & \(\checkmark\) & \textbf{0.897} & \textbf{0.928} & \textbf{0.970} & \textbf{0.976} & \textbf{0.899} & \textbf{0.940} \\
        \midrule
        Multi-block & \(\times\)     & 0.890 & 0.925 & 0.961 & 0.977 & 0.832 & 0.904 \\
        Multi-block & \(\checkmark\) & \textbf{0.903} & \textbf{0.934} & \textbf{0.973} & \textbf{0.980} & \textbf{0.908} & \textbf{0.938} \\
        \bottomrule
    \end{tabular}
\end{table*}

\section{Conclusion}
We presented ECG-JEPA, a self-supervised framework for 12-lead ECG representation learning whose design is tailored to the structure of multi-lead ECG signals. In particular, CroPA introduces a clinically motivated inductive bias for modeling cross-lead and temporal relationships, allowing the model to better reflect how ECGs are interpreted in practice.

Across \textit{PTB-XL}, \textit{CPSC2018}, and \textit{G12EC}, ECG-JEPA showed strong performance in diagnostic classification under both linear evaluation and fine-tuning. Beyond classification, the additional experiments on ECG feature extraction, ECG segmentation, and noise robustness indicate that the learned representations preserve clinically meaningful waveform structure and transfer effectively across diverse downstream tasks. These findings suggest that ECG-JEPA learns general-purpose ECG representations rather than features specialized to a single benchmark, supporting its potential as a foundation-model framework for ECG analysis.

\section*{Acknowledgments}
This work was supported by the National Research Foundation of Korea (NRF) Grant 2023R1A2C1005562, funded by the Korean government (MSIP). This research was also supported by the Bio\&Medical Technology Development Program of the National Research Foundation (NRF) funded by the Korean government (MSIT) (No. RS-2023-00222838). The author would like to thank Otto van Koert for his introduction to electrocardiography and for providing valuable insights on the subject.

\appendix

\section{Additional Experiments}

\subsection{Pretraining on a Larger ECG Dataset}\label{sec:extended pretraining}

Recent advances in machine learning have demonstrated that model performance often follows predictable scaling laws: as the size of the model and/or dataset increases, performance typically improves following a power-law relationship \citep{kaplan2020scaling}. This observation, originally reported in the context of natural language processing and computer vision tasks, motivates the investigation of whether similar trends hold in the ECG domain.

In our work, we sought to explore the impact of incorporating a larger dataset into ECG pretraining. However, the availability of large, open ECG datasets remains limited. With the exception of the MIMIC-IV-ECG v1.0 dataset—which initially contains approximately 800,000 12-lead, 10-second ECGs—most publicly available ECG datasets are relatively small. After excluding roughly 20,000 ECGs with missing values, about 780,000 samples were used in pretraining. Although MIMIC-IV-ECG offers a substantial amount of data, many of these ECGs are obtained from hospital admissions, emergency departments, and intensive care units, implying a bias towards more acute or critical conditions.

To evaluate the effect of this larger dataset on pretraining performance, we incorporated the MIMIC-IV-ECG dataset into our pretraining pipeline alongside our original datasets (Chapman, Ningbo, and CODE-15), and then assessed the resulting model on downstream tasks using the \textit{PTB-XL} and \textit{CPSC2018} datasets. In this preliminary study, we added the larger dataset all at once rather than incrementally, which may limit the precision of our analysis of scaling effects. Additionally, we did not increase the model size due to limited computational resources.

Tables \ref{tab:mimic_pretrain_linear} and \ref{tab:mimic_pretrain_finetune} provide comparisons of ECG-JEPA's performance when pretrained on either the original datasets or an extended dataset that includes the extensive MIMIC-IV-ECG collection. Notably, linear evaluation results show a slight drop in performance with the extended pretraining dataset, whereas fine-tuning performance remains very similar between the two settings. Despite the shift in pretraining data distribution, our findings indicate that the inclusion of MIMIC-IV-ECG data does not significantly degrade downstream performance, suggesting that the model is capable of extracting robust features even when trained on large datasets with inherent biases. These preliminary results warrant further investigation to fully understand the underlying dynamics and to optimize pretraining strategies for ECG data.

\begin{table*}[t]
    \centering
    \caption{\textbf{Pretraining-scale effect in linear evaluation}. Results for ECG-JEPA pretrained on 180k vs 960k samples.}
    \label{tab:mimic_pretrain_linear}
    \footnotesize
    \setlength{\tabcolsep}{4pt}
    \renewcommand{\arraystretch}{1.08}
    \begin{adjustbox}{max width=\textwidth}
    \begin{tabular}{llcccccccc}
        \toprule
        \multirow{2}{*}{Method} & \multirow{2}{*}{Pretrain Size} & \multicolumn{4}{c}{\textbf{Multi-label Task}} & \multicolumn{4}{c}{\textbf{Multi-class Task}} \\
        \cmidrule(lr){3-6}\cmidrule(lr){7-10}
        & & \multicolumn{2}{c}{\textit{PTB-XL}} & \multicolumn{2}{c}{\textit{CPSC2018}} & \multicolumn{2}{c}{\textit{PTB-XL}} & \multicolumn{2}{c}{\textit{CPSC2018}} \\
        \cmidrule(lr){3-4}\cmidrule(lr){5-6}\cmidrule(lr){7-8}\cmidrule(lr){9-10}
        & & AUC & F1 & AUC & F1 & AUC & F1 & AUC & F1 \\
        \midrule
        ECG-JEPA\(_{rb}\) (Original) & 180k & 0.906 & 0.665 & 0.965 & 0.767 & 0.897 & 0.640 & 0.970 & 0.793 \\
        ECG-JEPA\(_{rb}\) (Extended) & 960k & 0.900 & 0.675 & 0.969 & 0.779 & 0.879 & 0.613 & 0.970 & 0.772 \\
        \bottomrule
    \end{tabular}
    \end{adjustbox}
\end{table*}

\begin{table}[t]
    \centering
    \caption{\textbf{Pretraining-scale effect in fine-tuning}. Multi-class performance of ECG-JEPA after pretraining on 180k vs 960k samples.}
    \label{tab:mimic_pretrain_finetune}
    \footnotesize
    \setlength{\tabcolsep}{4pt}
    \renewcommand{\arraystretch}{1.08}
    \begin{tabular}{llcccc}
        \toprule
        \multirow{2}{*}{Method} & \multirow{2}{*}{Pretrain Size} & \multicolumn{2}{c}{\textit{PTB-XL}} & \multicolumn{2}{c}{\textit{CPSC2018}} \\
        \cmidrule(lr){3-4}\cmidrule(lr){5-6}
        & & AUC & F1 & AUC & F1 \\
        \midrule
        ECG-JEPA\(_{rb}\) (Original) & 180k & 0.938 & 0.691 & 0.978 & 0.843 \\
        ECG-JEPA\(_{rb}\) (Extended) & 960k & 0.936 & 0.721 & 0.977 & 0.826 \\
        \bottomrule
    \end{tabular}
\end{table}

\subsection{Effect of Learning Rates on Downstream Tasks} \label{app:lrs_performance}

We evaluate the sensitivity of ECG-JEPA to different base learning rates during downstream tasks—namely, linear evaluation and fine-tuning. Importantly, these experiments are conducted after pretraining; the learning rates analyzed here apply only to the downstream stages, not to pretraining.

\begin{figure*}[t!]
    \centering
    \includegraphics[width=\textwidth]{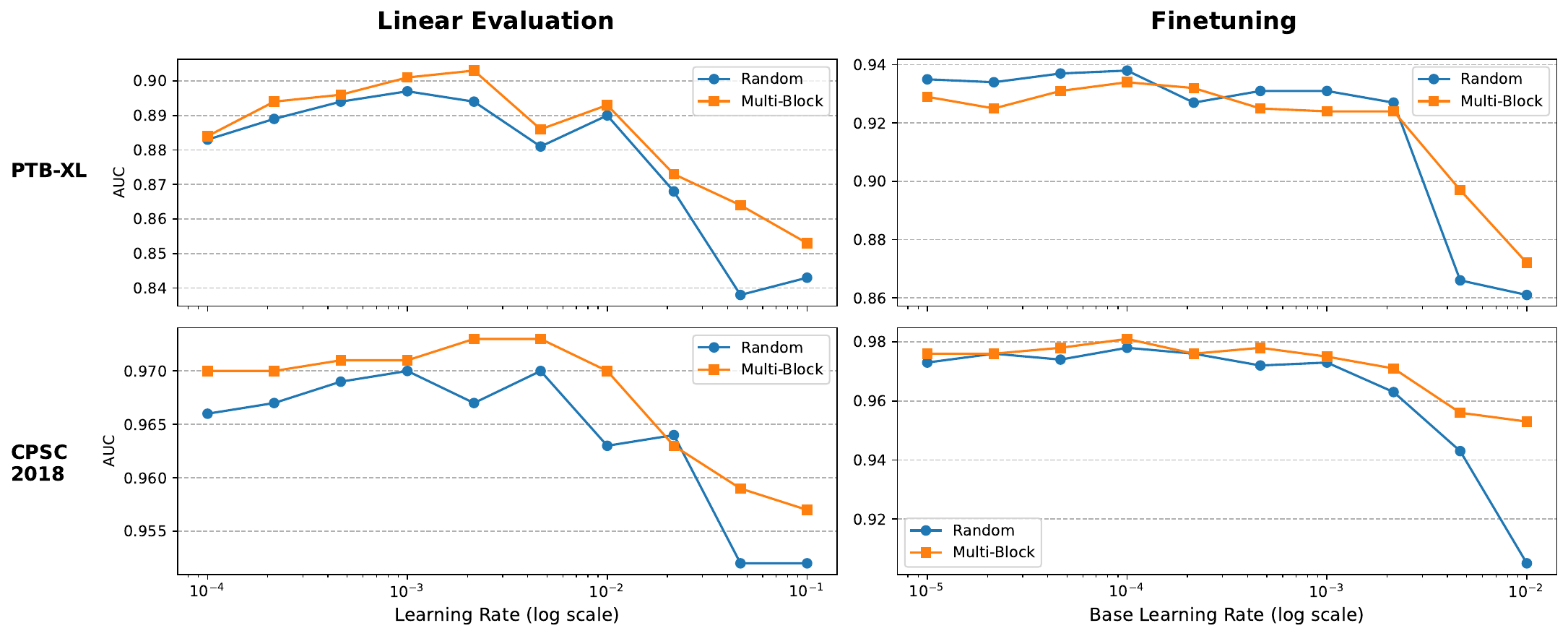}
    \caption{\textbf{Learning-rate sensitivity}. AUC across fine-tuning learning rates on \textit{PTB-XL} and \textit{CPSC2018} multi-class tasks.}
    \label{lrs}
\end{figure*}

For linear evaluation, we test 10 logarithmically spaced learning rates ranging from \(10^{-1}\) to \(10^{-4}\), while for fine-tuning, we test learning rates from \(10^{-2}\) to \(10^{-5}\). Figure~\ref{lrs} presents the AUC scores on the held-out test sets of \textit{PTB-XL} and \textit{CPSC2018} multi-class datasets.

Notably, ECG-JEPA achieves consistently high AUC across a wide range of learning rates for both evaluation modes. Although the optimal learning rate varies slightly across datasets and masking strategies, the overall robustness indicates that ECG-JEPA performs reliably without requiring extensive tuning. This makes ECG-JEPA practical for real-world clinical scenarios where computational constraints may limit exhaustive hyperparameter searches.

\subsection{Nearest Neighbor Classifier} \label{app:validation}
While linear probing and fine-tuning are common ways to evaluate SSL models, the \textit{Nearest Neighbor Classifier (NCC)} is a simple way to evaluate SSL models without further training. NCC is a lightweight multi-class classifier that does not involve training, and the class of a test sample is determined by the distribution of the training samples. Specifically, let \( \{ (x_{i}, y_{i}) \} \) be the set of pairs of training samples and labels. Assuming that there are \(C\) classes, we compute the class-mean vectors in the training set:
\[
\mu_{c} = \frac{1}{|\{y_{i}=c\}|} \sum_{y_{i} = c} x_{i}, \quad c \in [C].
\]
The test sample \(x\) is then classified according to the closest class-mean vectors, where the distance can be either Euclidean or cosine similarity. This method is similar to the \(k\)-nearest neighbor classifier, but it is simpler because it does not involve choice of \(k\). 

Table \ref{tab:ncc} shows NCC results across datasets. Note that we cannot compute AUC because NCC does not produce class probabilities. While ECG-JEPA still outperforms other methods, two points are worth noting. First, unlike linear evaluation, ECG-JEPA\(_{\text{rb}}\) performs better than ECG-JEPA\(_{\text{mb}}\). Second, the performance of SimCLR drops noticeably on \textit{CPSC2018}. While further analysis is needed to confirm the reason, we suspect that the class-mean vectors for SimCLR representations are less robust because per-class sample sizes are smaller in \textit{CPSC2018} than in \textit{PTB-XL}.

\begin{table}[t]
    \centering
    \caption{\textbf{Nearest-neighbor classifier}. Multi-class performance across \textit{PTB-XL} and \textit{CPSC2018}.}
    \label{tab:ncc}
    \scriptsize
    \setlength{\tabcolsep}{2.8pt}
    \renewcommand{\arraystretch}{1.05}
    \begin{tabular}{lcccccccc}
        \toprule
        \multirow{3}{*}{Method} & \multicolumn{4}{c}{Euclidean} & \multicolumn{4}{c}{Cosine} \\
        \cmidrule(lr){2-5}\cmidrule(lr){6-9}
        & \multicolumn{2}{c}{\textit{PTB-XL}} & \multicolumn{2}{c}{\textit{CPSC2018}} & \multicolumn{2}{c}{\textit{PTB-XL}} & \multicolumn{2}{c}{\textit{CPSC2018}} \\
        \cmidrule(lr){2-3}\cmidrule(lr){4-5}\cmidrule(lr){6-7}\cmidrule(lr){8-9}
        & Acc. & F1 & Acc. & F1 & Acc. & F1 & Acc. & F1 \\
        \midrule
        ST-MEM            & 0.524 & 0.419 & 0.611 & 0.571 & 0.524 & 0.420 & 0.613 & 0.574 \\
        SimCLR            & 0.567 & 0.452 & 0.498 & 0.443 & 0.560 & 0.445 & 0.497 & 0.440 \\
        ECG-FM            & 0.571 & 0.456 & 0.643 & 0.597 & 0.571 & 0.456 & 0.641 & 0.592 \\
        KED               & \underline{0.598} & \underline{0.461} & 0.623 & 0.552 & \underline{0.602} & \underline{0.460} & 0.628 & 0.567 \\
        ECG-JEPA\(_{rb}\) & \textbf{0.609} & \textbf{0.489} & \textbf{0.707} & \textbf{0.675} & \textbf{0.604} & \textbf{0.484} & \textbf{0.705} & \textbf{0.670} \\
        ECG-JEPA\(_{mb}\) & 0.584 & 0.446 & \underline{0.676} & \underline{0.644} & 0.577 & 0.444 & \underline{0.672} & \underline{0.640} \\
        \bottomrule
    \end{tabular}
\end{table}

\subsection{CroPA’s Effect: Statistical Significance Analysis}\label{statistical analysis on cropa}

To rigorously assess the impact of CroPA, we perform a statistical significance test using linear probes. We compare two pretrained models of ECG-JEPA\(_{rb}\): one pretrained with CroPA and the other without CroPA. We bootstrap the differences in AUC,
\[
\Delta \text{AUC} = \text{AUC}_{\text{CroPA}} - \text{AUC}_{\text{noCroPA}},
\]
by randomly sampling (with replacement) from the test set of each dataset, using a sample size equal to that of the full test set. This process is repeated 2000 times to compute the 95\% confidence interval of \(\Delta \text{AUC}\).

\begin{figure*}[t!]
    \centering
    \includegraphics[scale=0.5]{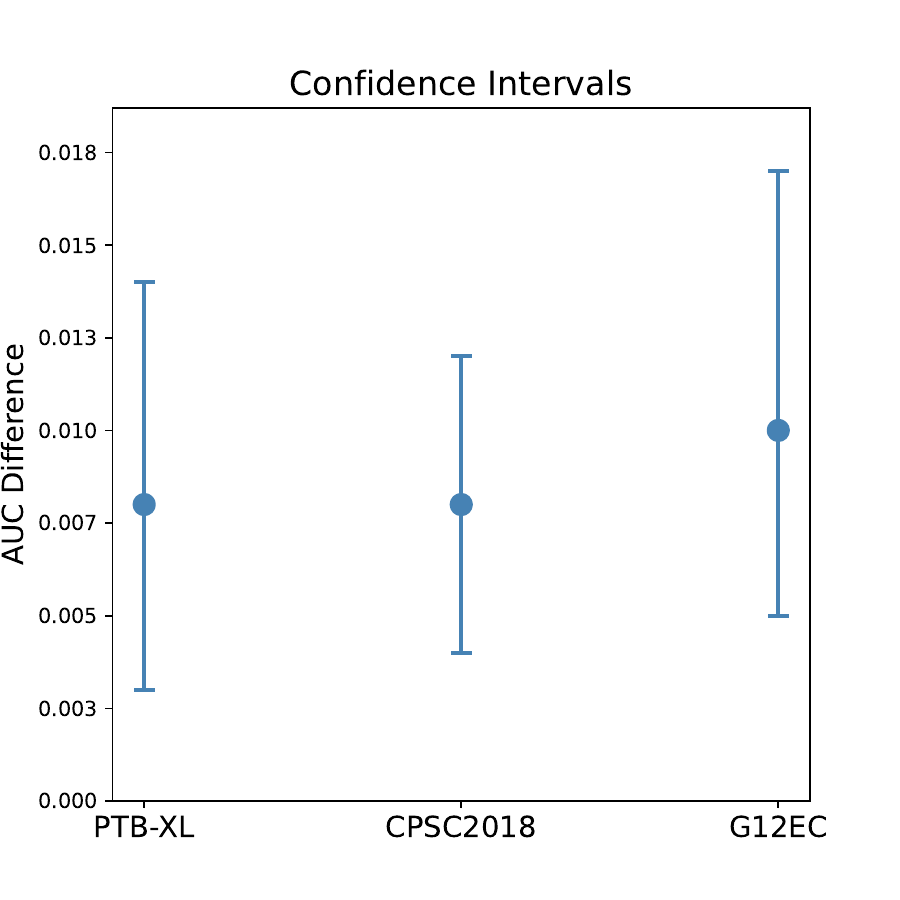}
    \caption{\textbf{CroPA confidence intervals}. Bootstrapped 95\% confidence intervals for $\Delta$AUC (CroPA - no CroPA) on \textit{PTB-XL}, \textit{CPSC2018}, and \textit{G12EC}; strictly positive intervals indicate statistically significant gains.}
    \label{fig:confidence_interval}
\end{figure*}

As shown in Figure \ref{fig:confidence_interval}, the bootstrapped 95\% confidence intervals for the AUC differences ($\Delta$AUC) between models pretrained with and without CroPA indicate a statistically significant improvement in performance. Although the improvement is modest, the strictly positive confidence intervals across all three datasets support the conclusion that CroPA yields a statistically significant improvement in performance.

\subsection{Visualization of ECG Representations}\label{sec:visualization}
Dimensionality reduction techniques enable the visualization of high-dimensional datasets, providing valuable insights into uncovering hidden patterns within complex data. UMAP \citep{umap}, a widely used non-linear dimensionality reduction method, balances local versus global structure in the data. 

In this section, we employ UMAP to visualize two prominent rhythm categories from \textit{PTB-XL}:  normal sinus rhythm (NSR) and atrial fibrillation (AFib). These labels comprise 16,687 samples (train: 15,021; test: 1,666) and 1,514 samples (train: 1,335; test: 149) in the rhythm category, respectively. See \ref{sec:dataset_details} for further explanation on the dataset. SR is characterized by a regular rhythm and a single P wave for each QRS complex, whereas AFib is characterized by irregular and often rapid heart rhythms. Although AFib is not directly related to diagnostic statements of the heart, it significantly increases the risk of stroke, heart failure, and other cardiovascular complications. 

\begin{figure*}
    \centering
    \includegraphics[scale=0.35]{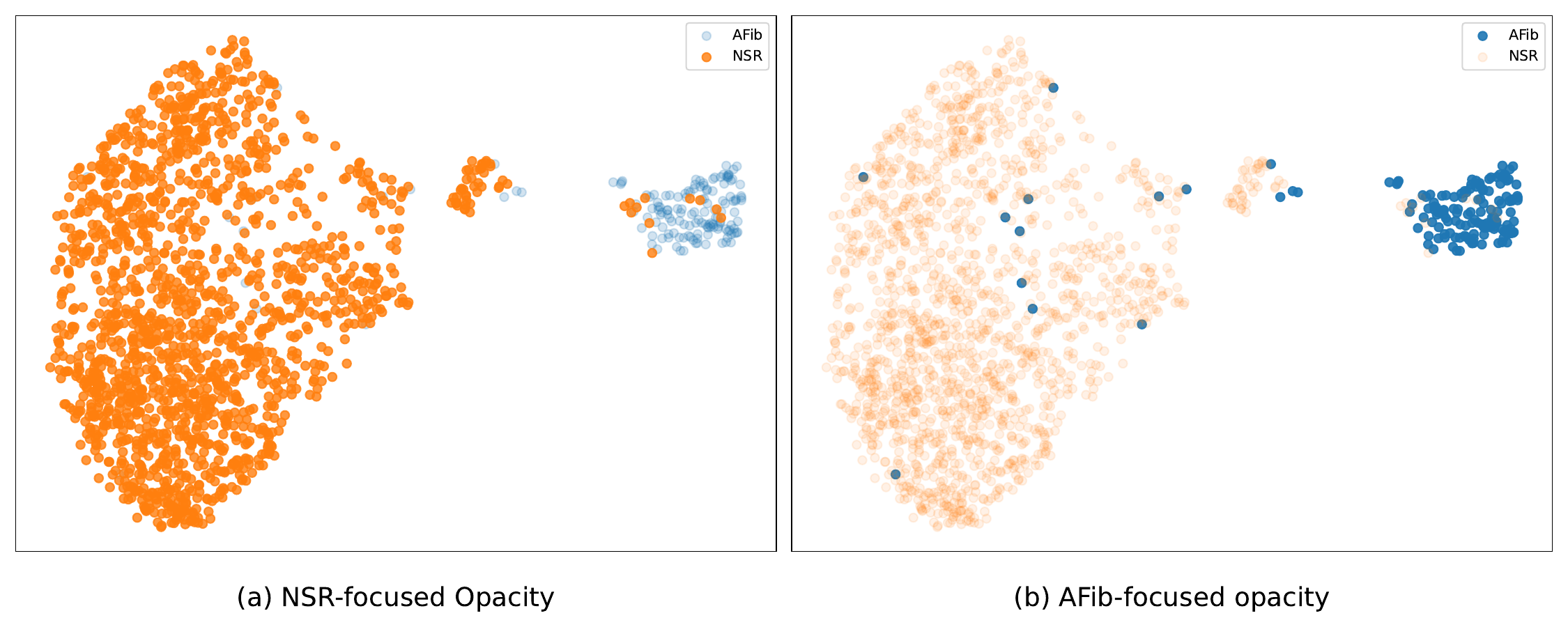}
    \caption{\textbf{UMAP of ECG representations}. NSR and AFib samples from the \textit{PTB-XL} test set.}
    \label{fig:afibumap}
\end{figure*}

Figure \ref{fig:afibumap} illustrates the UMAP projection of NSR and AFib samples from the test set, where UMAP is fitted on NSR and AFib samples from the training set. The majority of NSR ECGs (orange) and AFib ECGs (blue) are well-separated in the 2D space, though a few samples overlap with different clusters. These patterns highlight the need for further exploratory data analysis to better understand the structure and quality of the dataset. Notably, overlapping samples or outliers in unexpected clusters may indicate mislabeled instances. Such cases are examined in detail in \ref{app:case_study} to identify opportunities for enhancing the dataset's quality. This analysis demonstrates the potential of our model to aid in refining large-scale clinical datasets by uncovering hidden data issues.

\section{Ablation Study}
\subsection{Masking Ratio}\label{sec:maskingratio}
Table \ref{tab7:masking_strategy} presents the performance of ECG-JEPA in linear evaluation with different masking ratios and strategies. The results indicate that the model benefits from a high masking ratio. Notably, multi-block masking is advantageous for linear evaluation, while random masking is more effective for fine-tuning, as indicated in Table \ref{tab4:multiclass finetune}. Although random masking with a ratio of (0.7, 0.8) showed slightly better performance in the \textit{PTB-XL} multi-label task, this was observed in a later ablation study conducted after the main experiments had already adopted a (0.6, 0.7) ratio for consistency across tasks.

\begin{table}[t]
    \centering
    \caption{\textbf{Masking-strategy ablation}. Linear-evaluation results on \textit{PTB-XL} multi-label classification across masking ratios and strategies.}
    \label{tab7:masking_strategy}
    \footnotesize
    \setlength{\tabcolsep}{4pt}
    \renewcommand{\arraystretch}{1.08}
    \begin{tabular}{lcccc}
        \toprule
        Mask & Ratio & Freq. & AUC & F1 \\
        \midrule
        Random      & (0.3, 0.4)      & 1 & 0.883             & 0.658 \\
        Random      & (0.4, 0.5)      & 1 & 0.905             & \underline{0.678} \\
        Random      & (0.5, 0.6)      & 1 & \underline{0.906} & 0.691 \\
        Random      & (0.6, 0.7)      & 1 & \underline{0.906} & 0.665 \\
        Random      & (0.7, 0.8)      & 1 & \textbf{0.909}    & \textbf{0.706} \\
        \midrule
        Multi-block & (0.10, 0.15)    & 4 & 0.901              & 0.682 \\
        Multi-block & (0.15, 0.20)    & 4 & \underline{0.904}  & \underline{0.697} \\
        Multi-block & (0.175, 0.225)  & 4 & \textbf{0.912}     & \textbf{0.704} \\
        \bottomrule
    \end{tabular}
\end{table}

\subsection{Comparison with 12-Lead Model} \label{sec:12leadcomparison}
We now investigate the practical sufficiency of using 8 leads for ECG-JEPA pretraining. To evaluate the impact of this reduction, we trained models using both 8 leads and 12 leads and compared their performance on linear evaluation for the \textit{PTB-XL} multi-label task.

Table \ref{tab:lead_comparison} presents the results of this comparison using ECG-JEPA\(_{rb}\). As expected, the performance difference between the 8-lead and 12-lead models is minimal, indicating that using 8 leads is sufficient for effective pretraining without significant loss of information.

\begin{table}[t]
    \centering
    \caption{\textbf{Lead-count comparison}. Linear-evaluation performance of 8-lead and 12-lead models on \textit{PTB-XL} multi-label classification.}
    \label{tab:lead_comparison}
    \footnotesize
    \setlength{\tabcolsep}{4pt}
    \renewcommand{\arraystretch}{1.08}
    \begin{tabular}{lccc}
        \toprule
        Model & Epochs & AUC & F1 \\
        \midrule
        8-Lead  & 100 & 0.906 & 0.686 \\
        12-Lead & 100 & 0.905 & 0.699 \\
        \bottomrule
    \end{tabular}
\end{table}

\section{Experimental Details}\label{sec:Experimental Details}
\subsection{Downstream Datasets Details}\label{sec:dataset_details}
\begin{table*}[t]
    \centering
    \caption{\textbf{Downstream dataset split summary}. Number of ECG samples and label-space size used in multi-label and multi-class settings across \textit{PTB-XL}, \textit{CPSC2018}, and \textit{G12EC}.}
    \label{tab:dataset dist}
    \footnotesize
    \setlength{\tabcolsep}{4pt}
    \renewcommand{\arraystretch}{1.08}
    \begin{tabular}{llrrrrr}
        \toprule
        Dataset & Task & \# Classes & \# ECG (Total) & Train & Validation & Test \\
        \midrule
        \multirow{2}{*}{\textit{PTB-XL}}
            & Multi-label & 5  & 21388 & 17084 & 2146 & 2158 \\
            & Multi-class & 5  & 16244 & 12957 & 1637 & 1650 \\
        \midrule
        \multirow{2}{*}{\textit{CPSC2018}}
            & Multi-label & 9  & 9364  & 6554  & 937  & 1873 \\
            & Multi-class & 9  & 8682  & 6076  & 875  & 1731 \\
        \midrule
        \multirow{2}{*}{\textit{G12EC}}
            & Multi-label & 21 & 9412  & 7529  & 941  & 942 \\
            & Multi-class & 15 & 5101  & 4077  & 512  & 512 \\
        \bottomrule
    \end{tabular}
\end{table*}

\begin{table*}[t]
    \centering
    \caption{\textbf{\textit{PTB-XL} rhythm distribution}. Sample counts by split for multi-label and multi-class settings.}
    \label{tab:ptbxl rhythm dist}
    \footnotesize
    \setlength{\tabcolsep}{4pt}
    \renewcommand{\arraystretch}{1.08}
    \begin{tabular}{llrrrr}
        \toprule
        Type & Set & \# ECG & NSR & AFib & Others \\
        \midrule
        \multirow{3}{*}{Multi-label}
            & Total & 21030 & 16748 & 1514 & 2912 \\
            & Train & 18932 & 15074 & 1362 & 2625 \\
            & Test  & 2098  & 1674  & 152  & 287  \\
        \midrule
        \multirow{3}{*}{Multi-class}
            & Total & 20887 & 16687 & 1484 & 2716 \\
            & Train & 18804 & 15021 & 1335 & 2448 \\
            & Test  & 2083  & 1666  & 149  & 268  \\
        \bottomrule
    \end{tabular}
\end{table*}

Table \ref{tab:dataset dist} summarizes the dataset splits and label-space sizes for \textit{PTB-XL}, \textit{CPSC2018}, and \textit{G12EC}. Table \ref{tab:ptbxl rhythm dist} reports the rhythm-label distribution for \textit{PTB-XL}. Note that in multi-label settings, per-class sample counts can exceed the number of ECG recordings because each sample may have multiple labels.

The \textit{PTB-XL} dataset is stratified into ten folds. In our classification experiments, we use folds 1--8 for training, fold 9 for validation, and fold 10 for testing.

For the \textit{CPSC2018} dataset, only the training set is publicly available and is stratified into seven folds. In our split protocol, the first five folds are used for training, fold 6 for validation, and fold 7 for testing. The original \textit{CPSC2018} dataset consists of 6,877 ECG recordings, but we excluded recordings with a length of less than 10 seconds, resulting in 6,867 ECG recordings. Some recordings exceed 10 seconds in length; we subdivided these into non-overlapping 10-second segments and treated each as a separate sample, following the approach of \citet{guiding}.

For \textit{G12EC}, label construction follows a fixed, criterion-based pipeline. We start from the 27 scored SNOMED CT labels in the challenge and apply the official code-equivalence mapping defined by the PhysioNet/CinC 2020 challenge~\citep{cinc2020} (3 secondary codes merged into primary codes), resulting in 24 canonical labels. We then exclude recordings without any scored label and remove classes with fewer than 50 samples in the full dataset (that is, across the combined train, validation, and test pool); this removes 3 classes and leaves 21 labels for multi-label tasks. We then apply a reproducible 80/10/10 split (train/validation/test) saved in \texttt{g12ec\_split.json}. For multi-class tasks, we keep only recordings with exactly one label and again remove classes with fewer than 50 samples across the combined train, validation, and test pool, yielding 15 classes.

\subsection{Pretraining Hyperparameters}
Hyperparameters for ECG-JEPA pretraining are provided in Table \ref{tab:hyper_pretraining_rb}. In ECG-JEPA\(_{mb}\), the number of visible patches varies more than in ECG-JEPA\(_{rb}\), resulting in higher GPU memory usage. Consequently, we reduced the batch size to 64 to fit the model on a single NVIDIA RTX 3090 GPU. Interestingly, ECG-JEPA\(_{mb}\) benefits from larger learning rates, even with the halved batch size.

\begin{table}[t]
    \centering
    \caption{\textbf{Pretraining hyperparameters}. Settings for ECG-JEPA\(_{rb}\) and ECG-JEPA\(_{mb}\).}
    \label{tab:hyper_pretraining_rb}
    \footnotesize
    \setlength{\tabcolsep}{4pt}
    \renewcommand{\arraystretch}{1.08}
    \begin{tabular}{lll}
        \toprule
        Config & ECG-JEPA\(_{rb}\) & ECG-JEPA\(_{mb}\) \\
        \midrule
        Optimizer              & AdamW        & AdamW \\
        Learning rate          & 2.5e-5       & 5e-5 \\
        Weight decay           & 0.05         & 0.05 \\
        Batch size             & 128          & 64 \\
        Learning rate schedule & Cosine decay & Cosine decay \\
        Warmup epochs          & 5            & 5 \\
        Epochs                 & 100          & 100 \\
        Drop path              & 0.1          & 0.1 \\
        \bottomrule
    \end{tabular}
\end{table}

Besides ECG-JEPA, the additional baselines evaluated directly in this work are SimCLR, ST-MEM, ECG-FM, and KED \citep{simclr, guiding, mckeen2025ecg, tian2024foundation}. SimCLR uses a ResNet50 encoder \citep{resnet} (output dimension 2048) and was pretrained for 300 epochs; the checkpoint was selected from epochs 100, 200, and 300 based on \textit{PTB-XL} multi-label linear-evaluation performance.

Given SimCLR's sensitivity to augmentations, we used baseline shift, baseline wander, Gaussian noise, powerline noise (50\(\,\mathrm{Hz}\)), channel resize, random crop, and jump noise. For ST-MEM, ECG-FM, and KED, we used publicly available pretrained checkpoints. MoCo v3, MTAE, and MLAE are not retrained in our pipeline; their scores are reported from \citet{guiding}.

\subsection{Downstream Hyperparameters}
Table \ref{tab:hyper_downstream} summarizes the downstream settings for linear evaluation and fine-tuning. These settings are used for all model evaluations (ECG-JEPA, SimCLR, ST-MEM, ECG-FM, and KED), while model-specific input preprocessing is applied where required (e.g., ECG-FM: first 5s with per-sample z-score normalization; KED: resampling to 100 Hz).

For fine-tuning, the actual learning rate is calculated as \(lr = base\text{\_}lr \times batchsize / 256\), following the heuristic by \citet{baselr}.

\begin{table}[t]
    \centering
    \caption{\textbf{Downstream hyperparameters}. Settings used for linear evaluation and fine-tuning across all evaluated models.}
    \label{tab:hyper_downstream}
    \footnotesize
    \setlength{\tabcolsep}{4pt}
    \renewcommand{\arraystretch}{1.08}
    \begin{tabular}{lll}
        \toprule
        Config & Linear evaluation & Fine-tuning \\
        \midrule
        Optimizer                & AdamW & AdamW \\
        Learning-rate sweep      & \(10^{-1} \sim 10^{-4}\) & Base LR \(10^{-2} \sim 10^{-5}\) \\
        Weight decay             & 0.05 & 0.05 \\
        Batch size               & 32 & 16 \\
        LR schedule              & Cosine decay & Cosine decay \\
        Warmup epochs            & 3 & 3 \\
        Max epochs               & 100 & 100 \\
        Early stopping patience  & 10 & 10 \\
        \bottomrule
    \end{tabular}
\end{table}

\subsection{Noise Generation and Preprocessing for ECG Signals} \label{app:noise}
To evaluate robustness under noise (Section~\ref{exp:robustness}), we preprocess ECGs to generate noise-reduced data and add artificial noise to generate more severely corrupted inputs. Specifically, we apply high-pass and low-pass filters with cutoff frequencies 0.67\(\,\mathrm{Hz}\) and 40\(\,\mathrm{Hz}\), respectively. This removes most baseline drift and powerline interference.

While filtering is straightforward, realistic noise generation is less trivial. Following \citet{baselinenoise}, we model baseline drift as
\[
b(t) = C \cdot \sum_{k=0}^{K} a_k \cdot \cos\left(2\pi \cdot k \cdot \Delta f \cdot t + \phi_k\right)
\]
with \(\Delta f = f_{s}/N = 0.1 \, \mathrm{Hz}\), where \(f_s = 250 \, \mathrm{Hz}\) is the sampling frequency and \(N = 2500\) is the number of time steps. We use \(K = 5\), sample each amplitude coefficient \(a_k\) uniformly from \([0, 1]\), sample each phase \(\phi_k\) uniformly from \([0, 2\pi)\), and set the scaling factor to \(C = 0.5\).

For powerline interference, following \citet{powerlinenoise}, we use
\[
s(t) = C \cdot \sum_{k=1}^{K} a_k \cdot \cos\left(2\pi k f_n t + \phi\right),
\]
where \(f_n = 50 \, \mathrm{Hz}\) is the base powerline frequency, \(f_s = 250 \, \mathrm{Hz}\), \(N = 2500\), \(K = 3\), \(a_k \sim \mathcal{U}(0, 1)\), \(\phi \sim \mathcal{U}(0, 2\pi)\), and \(C = 0.5\).

\begin{figure*}[b]
    \centering
    \includegraphics[width=\textwidth]{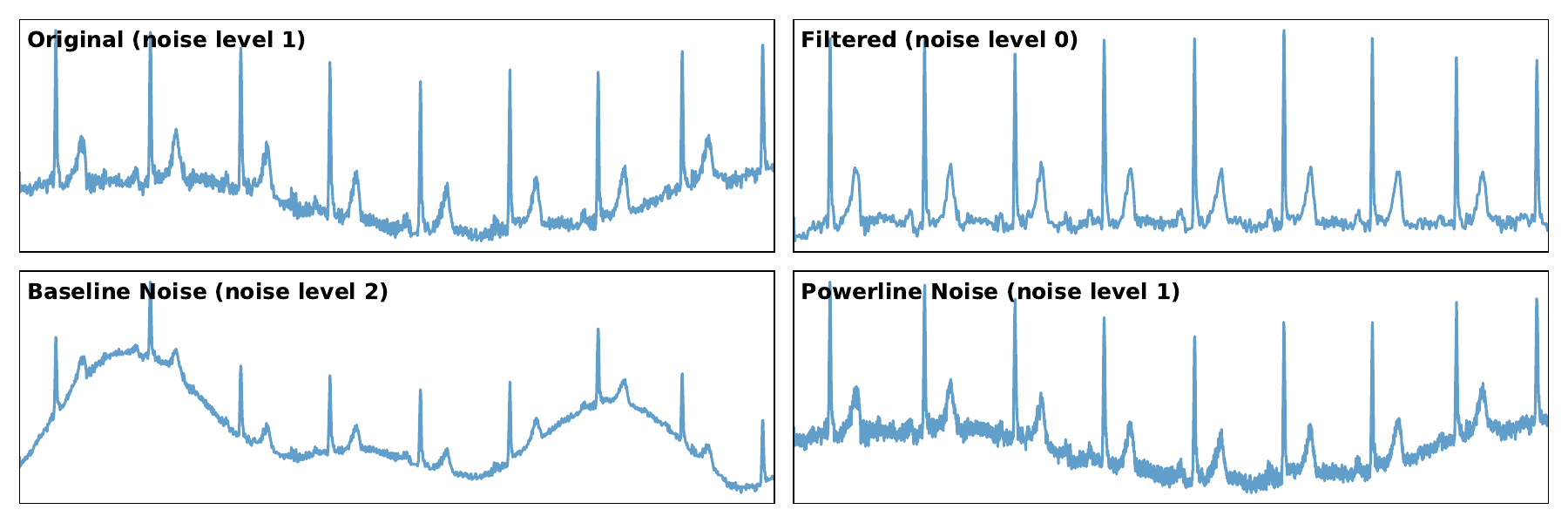}
    \caption{\textbf{Noise preprocessing examples}. Signals with filtering and added noise; the original signal contains mild baseline and powerline noise.}
    \label{noise}
\end{figure*}

Both noise types are applied to training and test ECGs with probability 0.5, and the same noise realization is added across all 8 leads. Figure~\ref{noise} illustrates the effects of filtering and added noise on representative ECG signals.

\subsection{Software Used in the Experiments}
All experiments were conducted using Python 3.10 on an Ubuntu 20.04 operating system. The primary framework utilized was PyTorch 2.3 for model implementation and training, with CUDA 11.8 for GPU acceleration.

\section{Exponential Moving Average}\label{sec:EMA}
The teacher network is initialized as a copy of the student network and is updated using an exponential moving average (EMA) of the student's weights. The EMA is computed as follows:
\[
\theta_{\text{teacher}}^i = \beta_i \theta_{\text{teacher}}^{i-1} + (1 - \beta_i) \theta_{\text{student}}^i
\]
where \(i\) denotes the current training iteration, and \(\beta_i\) is a momentum parameter that evolves during training. The momentum parameter \(\beta_i\) is computed as:
\[
\beta_i = \text{ema}_0 + \frac{i \cdot (\text{ema}_1 - \text{ema}_0)}{\text{iterations\_per\_epoch} \cdot \text{epochs}}
\]
Here, \(\text{ema}_0\) and \(\text{ema}_1\) represent the initial and final values of the momentum parameter, respectively. For our implementation, \(\text{ema}_0 = 0.996\) and \(\text{ema}_1 = 1.0\).

\section{Case Analysis of UMAP Embeddings} \label{app:case_study}
\begin{figure}[t]
    \centering
    \includegraphics[width=0.9\linewidth]{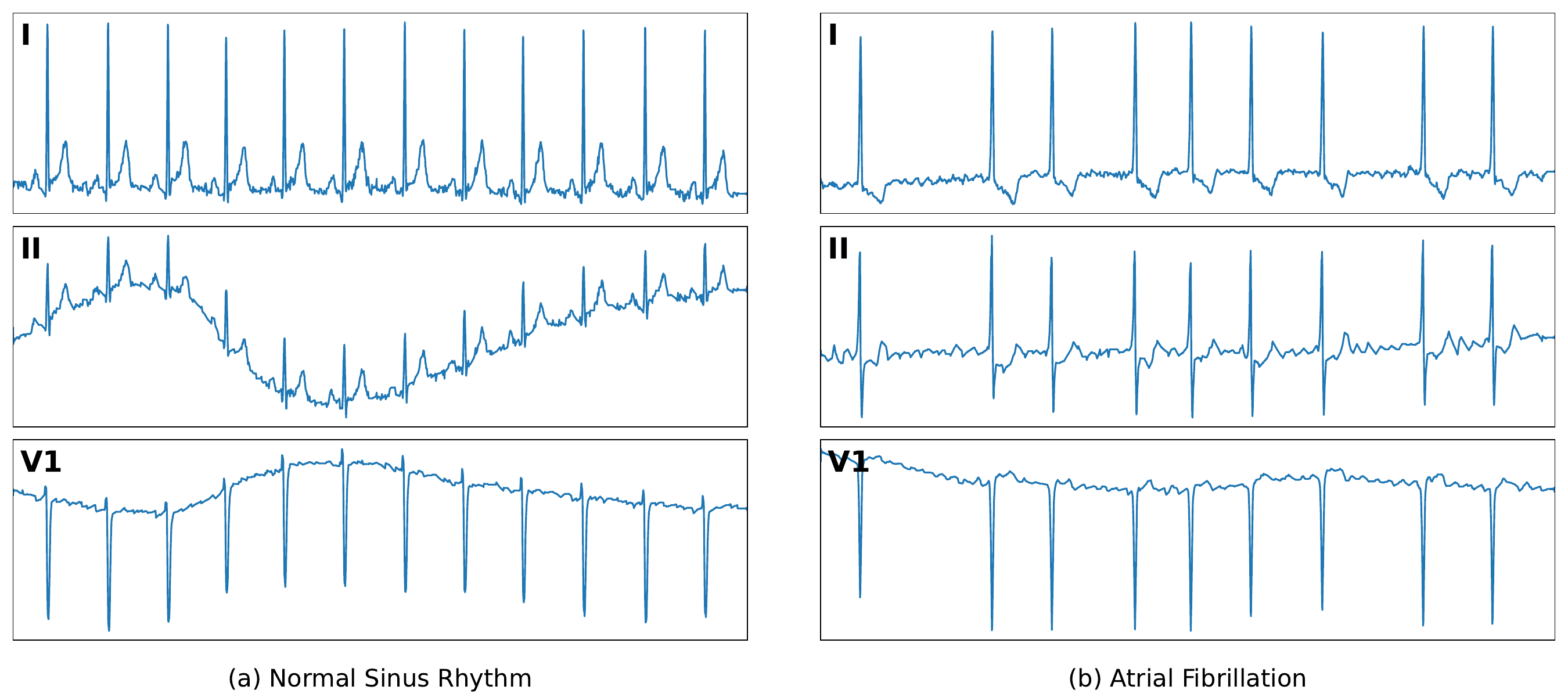}
    \caption{\textbf{NSR vs AFib examples}. Leads I, II, and V1; \textbf{(a)} NSR with regular rhythm and clear P waves, \textbf{(b)} AFib with irregular rhythm and absent P waves.}
    \label{fig:nsr_afib}
\end{figure}

In this section, we analyze individual ECG samples that are embedded in clusters different from their expected categories in the UMAP visualizations presented in Section \ref{sec:visualization}. These cases include normal sinus rhythm (NSR) samples located within atrial fibrillation (AFib) clusters and AFib samples found in NSR clusters. Such occurrences provide valuable insights into the model's learned representations and highlight the challenges posed by atypical or borderline samples.

NSR typically exhibits a regular heart rhythm with distinct P waves preceding each QRS complex. In contrast, AFib is characterized by an irregular rhythm, the absence of discernible P waves, and the presence of fibrillatory waves—irregular, rapid oscillations of the baseline. Figure \ref{fig:nsr_afib} consists of (a) an example of NSR and (b) an example of AFib, illustrating the characteristic differences between the two. However, certain samples in the UMAP embeddings deviate from these standard definitions. To further understand these cases, we review the ECG signals of selected samples from each scenario.

\subsection{NSR Samples in AFib Clusters}
\begin{figure}[h]
    \centering
    \includegraphics[width=\linewidth]{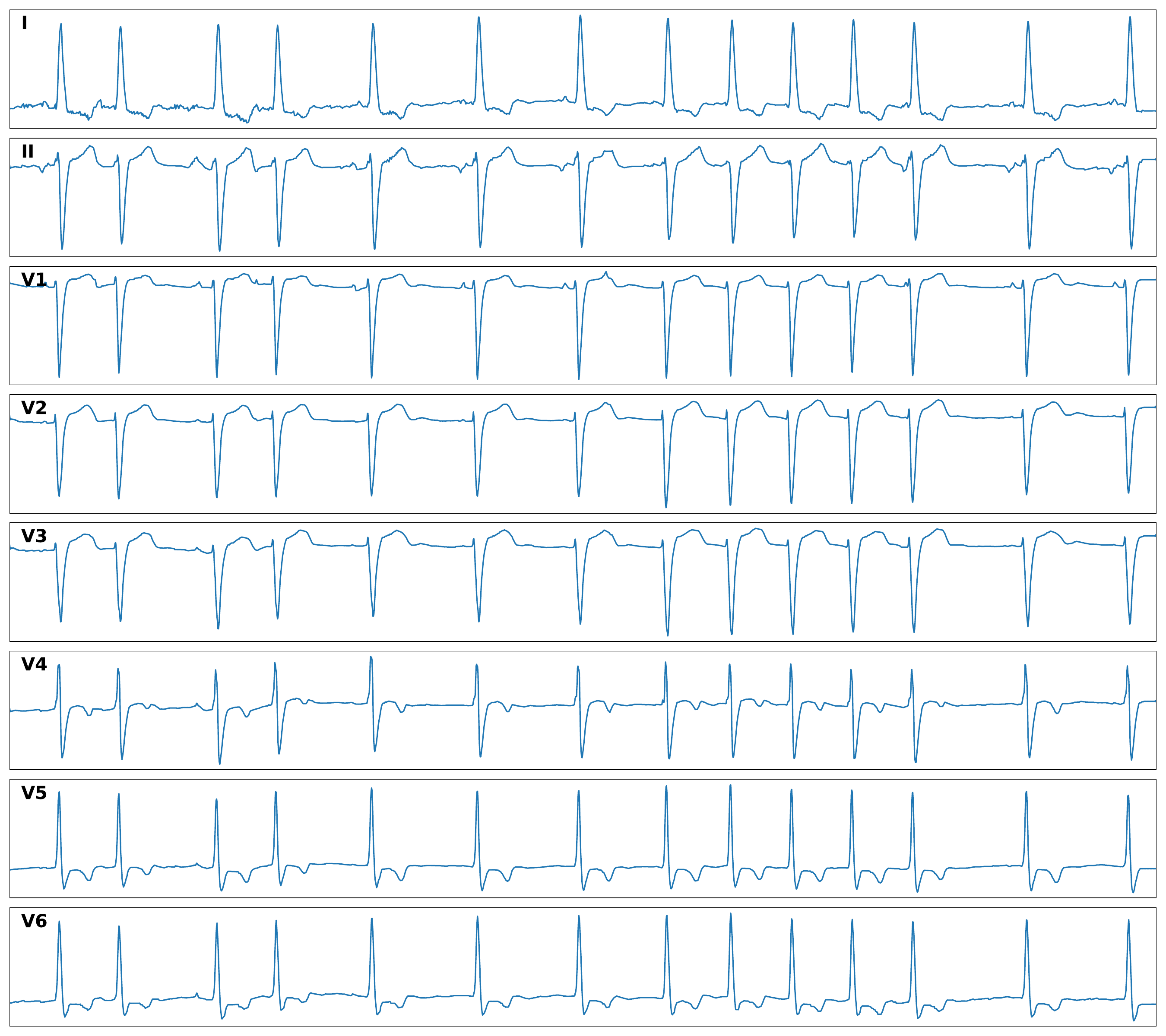}
    \caption{\textbf{NSR outlier example}. NSR sample embedded in the AFib cluster, showing irregular rhythm and missing P waves in leads V2-V6.}
    \label{fig:nsr_in_afib_cluster}
\end{figure}

Figure \ref{fig:nsr_in_afib_cluster} shows an example of an NSR signal that is embedded in the AFib cluster. Upon inspection, this signal reveals irregularities in rhythm, and P waves are missing in leads V2-V6. These features, while atypical for NSR, may explain why the model's representation aligns this signal with the AFib cluster.

\subsection{AFib Samples in NSR Clusters}
\begin{figure}[h]
    \centering
    \includegraphics[width=\linewidth]{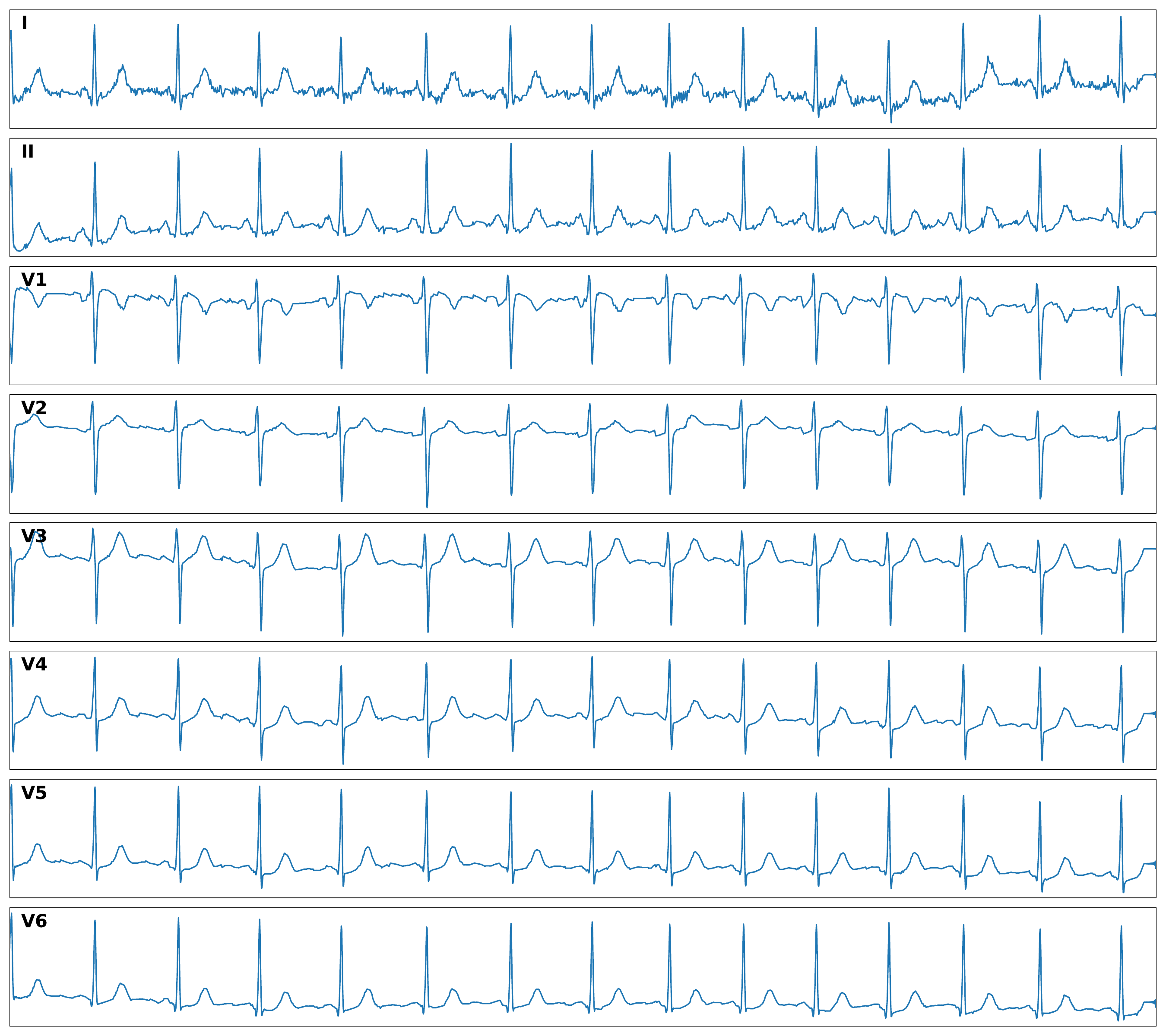}
    \caption{\textbf{AFib outlier example}. AFib sample embedded in the NSR cluster, showing irregular P waves with partially NSR-like rhythm.}
    \label{fig:afib_in_nsr_cluster}
\end{figure}

Conversely, Figure \ref{fig:afib_in_nsr_cluster} illustrates an AFib signal that is embedded in the NSR cluster. While this signal shows fibrillatory waves in leads I,II, and V1, the rhythm is regular and P waves are visible. This partial resemblance to NSR may have caused the model to assign it to the NSR cluster.

\subsection{Implications of Atypical Cases} 
The presence of these atypical cases underscores the complexity of real-world ECG classification. Such samples may reflect physiological conditions that do not strictly align with the standard definitions of NSR or AFib, highlighting the potential for borderline or transitional states. Additionally, these cases might indicate mislabeled data, which is not uncommon given the inherent complexity of ECG interpretation.

Our analysis demonstrates that the model's learned representations are valuable not only for classifying typical cases but also for identifying and interpreting atypical cases. By examining UMAP embeddings, the model provides insights into ambiguous samples and helps uncover potential labeling inconsistencies. This capability is particularly useful, as it can contribute to improving dataset quality by detecting and addressing mislabeled or borderline cases.

\bibliographystyle{unsrtnat}
\bibliography{references}

@misc{guiding,
  title={Guiding Masked Representation Learning to Capture Spatio-Temporal Relationship of Electrocardiogram},
  author={Na, Yeongyeon and Park, Minje and Tae, Yunwon and Joo, Sunghoon},
  year={2024},
  eprint={2402.09450},
  archivePrefix={arXiv},
  primaryClass={eess.SP}
}

@article{chapman,
  title={A 12-lead electrocardiogram database for arrhythmia research covering more than 10,000 patients},
  author={Zheng, Jianwei and Zhang, Jianming and Danioko, Sidy and Yao, Hai and Guo, Hangyuan and Rakovski, Cyril},
  journal={Scientific data},
  volume={7},
  number={1},
  pages={48},
  year={2020},
  publisher={Nature Publishing Group UK London}
}

@article{ningbo,
  title={Optimal multi-stage arrhythmia classification approach},
  author={Zheng, Jianwei and Chu, Huimin and Struppa, Daniele and Zhang, Jianming and Yacoub, Sir Magdi and El-Askary, Hesham and Chang, Anthony and Ehwerhemuepha, Louis and Abudayyeh, Islam and Barrett, Alexander and others},
  journal={Scientific reports},
  volume={10},
  number={1},
  pages={2898},
  year={2020},
  publisher={Nature Publishing Group UK London}
}

@inproceedings{chen2019large,
  title={Large-scale classification of 12-lead ECG with deep learning},
  author={Chen, Yu-Jhen and Liu, Chien-Liang and Tseng, Vincent S and Hu, Yu-Feng and Chen, Shih-Ann},
  booktitle={2019 IEEE EMBS international conference on biomedical \& health informatics (BHI)},
  pages={1--4},
  year={2019},
  organization={IEEE}
}

@article{ptbxl,
  title={PTB-XL, a large publicly available electrocardiography dataset},
  author={Wagner, Patrick and Strodthoff, Nils and Bousseljot, Ralf-Dieter and Kreiseler, Dieter and Lunze, Fatima I and Samek, Wojciech and Schaeffter, Tobias},
  journal={Scientific data},
  volume={7},
  number={1},
  pages={1--15},
  year={2020},
  publisher={Nature Publishing Group}
}

@article{cpsc,
  title={An open access database for evaluating the algorithms of electrocardiogram rhythm and morphology abnormality detection},
  author={Liu, Feifei and Liu, Chengyu and Zhao, Lina and Zhang, Xiangyu and Wu, Xiaoling and Xu, Xiaoyan and Liu, Yulin and Ma, Caiyun and Wei, Shoushui and He, Zhiqiang and others},
  journal={Journal of Medical Imaging and Health Informatics},
  volume={8},
  number={7},
  pages={1368--1373},
  year={2018},
  publisher={American Scientific Publishers}
}

@misc{bert,
      title={BERT: Pre-training of Deep Bidirectional Transformers for Language Understanding}, 
      author={Jacob Devlin and Ming-Wei Chang and Kenton Lee and Kristina Toutanova},
      year={2019},
      eprint={1810.04805},
      archivePrefix={arXiv},
      primaryClass={id='cs.CL' full_name='Computation and Language' is_active=True alt_name='cmp-lg' in_archive='cs' is_general=False description='Covers natural language processing. Roughly includes material in ACM Subject Class I.2.7. Note that work on artificial languages (programming languages, logics, formal systems) that does not explicitly address natural-language issues broadly construed (natural-language processing, computational linguistics, speech, text retrieval, etc.) is not appropriate for this area.'}
}

@misc{ssl_robust,
      title={Self-supervised Learning is More Robust to Dataset Imbalance}, 
      author={Hong Liu and Jeff Z. HaoChen and Adrien Gaidon and Tengyu Ma},
      year={2022},
      eprint={2110.05025},
      archivePrefix={arXiv},
      primaryClass={id='cs.LG' full_name='Machine Learning' is_active=True alt_name=None in_archive='cs' is_general=False description='Papers on all aspects of machine learning research (supervised, unsupervised, reinforcement learning, bandit problems, and so on) including also robustness, explanation, fairness, and methodology. cs.LG is also an appropriate primary category for applications of machine learning methods.'}
}

@article{gpt3,
  title={Language models are few-shot learners},
  author={Brown, Tom and Mann, Benjamin and Ryder, Nick and Subbiah, Melanie and Kaplan, Jared D and Dhariwal, Prafulla and Neelakantan, Arvind and Shyam, Pranav and Sastry, Girish and Askell, Amanda and others},
  journal={Advances in neural information processing systems},
  volume={33},
  pages={1877--1901},
  year={2020}
}

@misc{llama,
      title={LLaMA: Open and Efficient Foundation Language Models}, 
      author={Hugo Touvron and Thibaut Lavril and Gautier Izacard and Xavier Martinet and Marie-Anne Lachaux and Timothée Lacroix and Baptiste Rozière and Naman Goyal and Eric Hambro and Faisal Azhar and Aurelien Rodriguez and Armand Joulin and Edouard Grave and Guillaume Lample},
      year={2023},
      eprint={2302.13971},
      archivePrefix={arXiv},
      primaryClass={id='cs.CL' full_name='Computation and Language' is_active=True alt_name='cmp-lg' in_archive='cs' is_general=False description='Covers natural language processing. Roughly includes material in ACM Subject Class I.2.7. Note that work on artificial languages (programming languages, logics, formal systems) that does not explicitly address natural-language issues broadly construed (natural-language processing, computational linguistics, speech, text retrieval, etc.) is not appropriate for this area.'}
}

@inproceedings{simclr,
  title={A simple framework for contrastive learning of visual representations},
  author={Chen, Ting and Kornblith, Simon and Norouzi, Mohammad and Hinton, Geoffrey},
  booktitle={International conference on machine learning},
  pages={1597--1607},
  year={2020},
  organization={PMLR}
}

@article{byol,
  title={Bootstrap your own latent-a new approach to self-supervised learning},
  author={Grill, Jean-Bastien and Strub, Florian and Altch{\'e}, Florent and Tallec, Corentin and Richemond, Pierre and Buchatskaya, Elena and Doersch, Carl and Avila Pires, Bernardo and Guo, Zhaohan and Gheshlaghi Azar, Mohammad and others},
  journal={Advances in neural information processing systems},
  volume={33},
  pages={21271--21284},
  year={2020}
}

@inproceedings{mae,
  title={Masked autoencoders are scalable vision learners},
  author={He, Kaiming and Chen, Xinlei and Xie, Saining and Li, Yanghao and Doll{\'a}r, Piotr and Girshick, Ross},
  booktitle={Proceedings of the IEEE/CVF conference on computer vision and pattern recognition},
  pages={16000--16009},
  year={2022}
}

@inproceedings{ijepa,
  title={Self-supervised learning from images with a joint-embedding predictive architecture},
  author={Assran, Mahmoud and Duval, Quentin and Misra, Ishan and Bojanowski, Piotr and Vincent, Pascal and Rabbat, Michael and LeCun, Yann and Ballas, Nicolas},
  booktitle={Proceedings of the IEEE/CVF Conference on Computer Vision and Pattern Recognition},
  pages={15619--15629},
  year={2023}
}

@misc{vjepa,
      title={Revisiting Feature Prediction for Learning Visual Representations from Video}, 
      author={Adrien Bardes and Quentin Garrido and Jean Ponce and Xinlei Chen and Michael Rabbat and Yann LeCun and Mahmoud Assran and Nicolas Ballas},
      year={2024},
      eprint={2404.08471},
      archivePrefix={arXiv},
      primaryClass={id='cs.CV' full_name='Computer Vision and Pattern Recognition' is_active=True alt_name=None in_archive='cs' is_general=False description='Covers image processing, computer vision, pattern recognition, and scene understanding. Roughly includes material in ACM Subject Classes I.2.10, I.4, and I.5.'}
}

@article{vmae,
  title={Videomae: Masked autoencoders are data-efficient learners for self-supervised video pre-training},
  author={Tong, Zhan and Song, Yibing and Wang, Jue and Wang, Limin},
  journal={Advances in neural information processing systems},
  volume={35},
  pages={10078--10093},
  year={2022}
}

@inproceedings{clocs,
  title={Clocs: Contrastive learning of cardiac signals across space, time, and patients},
  author={Kiyasseh, Dani and Zhu, Tingting and Clifton, David A},
  booktitle={International Conference on Machine Learning},
  pages={5606--5615},
  year={2021},
  organization={PMLR}
}

@article{cpcecg,
  title={Self-supervised representation learning from 12-lead ECG data},
  author={Mehari, Temesgen and Strodthoff, Nils},
  journal={Computers in biology and medicine},
  volume={141},
  pages={105114},
  year={2022},
  publisher={Elsevier}
}

@article{maefe,
  title={Maefe: Masked autoencoders family of electrocardiogram for self-supervised pretraining and transfer learning},
  author={Zhang, Huaicheng and Liu, Wenhan and Shi, Jiguang and Chang, Sheng and Wang, Hao and He, Jin and Huang, Qijun},
  journal={IEEE Transactions on Instrumentation and Measurement},
  volume={72},
  pages={1--15},
  year={2022},
  publisher={IEEE}
}

@inproceedings{dae,
  title={Extracting and composing robust features with denoising autoencoders},
  author={Vincent, Pascal and Larochelle, Hugo and Bengio, Yoshua and Manzagol, Pierre-Antoine},
  booktitle={Proceedings of the 25th international conference on Machine learning},
  pages={1096--1103},
  year={2008}
}

@article{segmentation,
  title={Deep learning based ECG segmentation for delineation of diverse arrhythmias},
  author={Joung, Chankyu and Kim, Mijin and Paik, Taejin and Kong, Seong-Ho and Oh, Seung-Young and Jeon, Won Kyeong and Jeon, Jae-hu and Hong, Joong-Sik and Kim, Wan-Joong and Kook, Woong and others},
  journal={PloS one},
  volume={19},
  number={6},
  pages={e0303178},
  year={2024},
  publisher={Public Library of Science San Francisco, CA USA}
}

@misc{balestriero2024,
      title={Learning by Reconstruction Produces Uninformative Features For Perception}, 
      author={Randall Balestriero and Yann LeCun},
      year={2024},
      eprint={2402.11337},
      archivePrefix={arXiv},
      primaryClass={cs.CV},
      url={https://arxiv.org/abs/2402.11337}, 
}

@misc{cpc,
      title={Representation Learning with Contrastive Predictive Coding}, 
      author={Aaron van den Oord and Yazhe Li and Oriol Vinyals},
      year={2019},
      eprint={1807.03748},
      archivePrefix={arXiv},
      primaryClass={cs.LG},
      url={https://arxiv.org/abs/1807.03748}, 
}

@misc{path,
  title={A path towards autonomous machine intelligence version 0.9. 2, 2022-06-27},
  author={LeCun, Yann},
  journal={Open Review},
  volume={62},
  number={1},
  pages={1--62},
  year={2022},
  howpublished="\url{https://openreview.net/forum?id=BZ5a1r-kVsf}",
  note="Accessed: 2024-06-01"
}

@misc{baselr,
      title={Accurate, Large Minibatch SGD: Training ImageNet in 1 Hour}, 
      author={Priya Goyal and Piotr Dollár and Ross Girshick and Pieter Noordhuis and Lukasz Wesolowski and Aapo Kyrola and Andrew Tulloch and Yangqing Jia and Kaiming He},
      year={2018},
      eprint={1706.02677},
      archivePrefix={arXiv},
      primaryClass={cs.CV},
      url={https://arxiv.org/abs/1706.02677}, 
}

@inproceedings{resnet,
  title={Deep residual learning for image recognition},
  author={He, Kaiming and Zhang, Xiangyu and Ren, Shaoqing and Sun, Jian},
  booktitle={Proceedings of the IEEE conference on computer vision and pattern recognition},
  pages={770--778},
  year={2016}
}

@inproceedings{sereda2019ecg,
  title={ECG segmentation by neural networks: Errors and correction},
  author={Sereda, Iana and Alekseev, Sergey and Koneva, Aleksandra and Kataev, Roman and Osipov, Grigory},
  booktitle={2019 International Joint Conference on Neural Networks (IJCNN)},
  pages={1--7},
  year={2019},
  organization={IEEE}
}

@inproceedings{moskalenko2020deep,
  title={Deep learning for ECG segmentation},
  author={Moskalenko, Viktor and Zolotykh, Nikolai and Osipov, Grigory},
  booktitle={Advances in Neural Computation, Machine Learning, and Cognitive Research III: Selected Papers from the XXI International Conference on Neuroinformatics, October 7-11, 2019, Dolgoprudny, Moscow Region, Russia},
  pages={246--254},
  year={2020},
  organization={Springer}
}

@article{chen2023post,
  title={Post-processing refined ECG delineation based on 1D-UNet},
  author={Chen, Zhenqin and Wang, Mengying and Zhang, Meiyu and Huang, Wei and Gu, Hanjie and Xu, Jinshan},
  journal={Biomedical Signal Processing and Control},
  volume={79},
  pages={104106},
  year={2023},
  publisher={Elsevier}
}

@article{hannun2019cardiologist,
  title={Cardiologist-level arrhythmia detection and classification in ambulatory electrocardiograms using a deep neural network},
  author={Hannun, Awni Y and Rajpurkar, Pranav and Haghpanahi, Masoumeh and Tison, Geoffrey H and Bourn, Codie and Turakhia, Mintu P and Ng, Andrew Y},
  journal={Nature medicine},
  volume={25},
  number={1},
  pages={65--69},
  year={2019},
  publisher={Nature Publishing Group US New York}
}

@article{ribeiro2020automatic,
  title={Automatic diagnosis of the 12-lead ECG using a deep neural network},
  author={Ribeiro, Ant{\^o}nio H and Ribeiro, Manoel Horta and Paix{\~a}o, Gabriela MM and Oliveira, Derick M and Gomes, Paulo R and Canazart, J{\'e}ssica A and Ferreira, Milton PS and Andersson, Carl R and Macfarlane, Peter W and Meira Jr, Wagner and others},
  journal={Nature communications},
  volume={11},
  number={1},
  pages={1760},
  year={2020},
  publisher={Nature Publishing Group UK London}
}

@article{siontis2021artificial,
  title={Artificial intelligence-enhanced electrocardiography in cardiovascular disease management},
  author={Siontis, Konstantinos C and Noseworthy, Peter A and Attia, Zachi I and Friedman, Paul A},
  journal={Nature Reviews Cardiology},
  volume={18},
  number={7},
  pages={465--478},
  year={2021},
  publisher={Nature Publishing Group UK London}
}

@misc{cookbook,
      title={A Cookbook of Self-Supervised Learning}, 
      author={Randall Balestriero and Mark Ibrahim and Vlad Sobal and Ari Morcos and Shashank Shekhar and Tom Goldstein and Florian Bordes and Adrien Bardes and Gregoire Mialon and Yuandong Tian and Avi Schwarzschild and Andrew Gordon Wilson and Jonas Geiping and Quentin Garrido and Pierre Fernandez and Amir Bar and Hamed Pirsiavash and Yann LeCun and Micah Goldblum},
      year={2023},
      eprint={2304.12210},
      archivePrefix={arXiv},
      primaryClass={cs.LG},
      url={https://arxiv.org/abs/2304.12210}, 
}

@misc{vicreg,
      title={VICReg: Variance-Invariance-Covariance Regularization for Self-Supervised Learning}, 
      author={Adrien Bardes and Jean Ponce and Yann LeCun},
      year={2022},
      eprint={2105.04906},
      archivePrefix={arXiv},
      primaryClass={cs.CV},
      url={https://arxiv.org/abs/2105.04906}, 
}

@misc{simsiam,
      title={Exploring Simple Siamese Representation Learning}, 
      author={Xinlei Chen and Kaiming He},
      year={2020},
      eprint={2011.10566},
      archivePrefix={arXiv},
      primaryClass={cs.CV},
      url={https://arxiv.org/abs/2011.10566}, 
}

@book{thaler2021only,
  title={The only EKG book you’ll ever need},
  author={Thaler, Malcolm S},
  year={2021},
  publisher={Lippincott Williams \& Wilkins}
}

@article{baselinenoise,
  title={Comparison of baseline wander removal techniques considering the preservation of ST changes in the ischemic ECG: a simulation study},
  author={Lenis, Gustavo and Pilia, Nicolas and Loewe, Axel and Schulze, Walther HW and D{\"o}ssel, Olaf},
  journal={Computational and mathematical methods in medicine},
  volume={2017},
  number={1},
  pages={9295029},
  year={2017},
  publisher={Wiley Online Library}
}

@article{powerlinenoise,
  title={A comparison of the noise sensitivity of nine QRS detection algorithms},
  author={Friesen, Gary M and Jannett, Thomas C and Jadallah, Manal Afify and Yates, Stanford L and Quint, Stephen R and Nagle, H Troy},
  journal={IEEE Transactions on biomedical engineering},
  volume={37},
  number={1},
  pages={85--98},
  year={1990},
  publisher={IEEE}
}

@article{umap,
  title={Umap: Uniform manifold approximation and projection for dimension reduction},
  author={McInnes, Leland and Healy, John and Melville, James},
  journal={arXiv preprint arXiv:1802.03426},
  year={2018}
}

@article{cinc2020,
  title={Classification of 12-lead ecgs: the physionet/computing in cardiology challenge 2020},
  author={Alday, Erick A Perez and Gu, Annie and Shah, Amit J and Robichaux, Chad and Wong, An-Kwok Ian and Liu, Chengyu and Liu, Feifei and Rad, Ali Bahrami and Elola, Andoni and Seyedi, Salman and others},
  journal={Physiological measurement},
  volume={41},
  number={12},
  pages={124003},
  year={2020},
  publisher={IOP Publishing}
}

@article{hu2023spatiotemporal,
  title={Spatiotemporal self-supervised representation learning from multi-lead ECG signals},
  author={Hu, Rui and Chen, Jie and Zhou, Li},
  journal={Biomedical Signal Processing and Control},
  volume={84},
  pages={104772},
  year={2023},
  publisher={Elsevier}
}

@inproceedings{oh2022lead,
  title={Lead-agnostic self-supervised learning for local and global representations of electrocardiogram},
  author={Oh, Jungwoo and Chung, Hyunseung and Kwon, Joon-myoung and Hong, Dong-gyun and Choi, Edward},
  booktitle={Conference on Health, Inference, and Learning},
  pages={338--353},
  year={2022},
  organization={PMLR}
}

@article{wav2vec2,
  title={wav2vec 2.0: A framework for self-supervised learning of speech representations},
  author={Baevski, Alexei and Zhou, Yuhao and Mohamed, Abdelrahman and Auli, Michael},
  journal={Advances in neural information processing systems},
  volume={33},
  pages={12449--12460},
  year={2020}
}

@inproceedings{yang2022masked,
  title={Masked autoencoder for ECG representation learning},
  author={Yang, Shunxiang and Lian, Cheng and Zeng, Zhigang},
  booktitle={2022 12th International Conference on Information Science and Technology (ICIST)},
  pages={95--98},
  year={2022},
  organization={IEEE}
}

@inproceedings{wang2023unsupervised,
  title={Unsupervised pre-training using masked autoencoders for ecg analysis},
  author={Wang, Guoxin and Wang, Qingyuan and Iyer, Ganesh Neelakanta and Nag, Avishek and John, Deepu},
  booktitle={2023 IEEE Biomedical Circuits and Systems Conference (BioCAS)},
  pages={1--5},
  year={2023},
  organization={IEEE}
}

@article{mckeen2025ecg,
  title={ECG-FM: an open electrocardiogram foundation model},
  author={McKeen, Kaden and Masood, Sameer and Toma, Augustin and Rubin, Barry and Wang, Bo},
  journal={Jamia Open},
  volume={8},
  number={5},
  pages={ooaf122},
  year={2025},
  publisher={Oxford University Press}
}

@article{tian2024foundation,
  title={Foundation model of {ECG} diagnosis: Diagnostics and explanations of any form and rhythm on {ECG}},
  author={Tian, Yuanyuan and Li, Zhiyuan and Jin, Yanrui and Wang, Mengxiao and Wei, Xiaoyang and Zhao, Liqun and Liu, Yunqing and Liu, Jinlei and Liu, Chengliang},
  journal={Cell Reports Medicine},
  volume={5},
  number={12},
  pages={101875},
  year={2024},
  doi={10.1016/j.xcrm.2024.101875}
}

@article{kaplan2020scaling,
  title={Scaling laws for neural language models},
  author={Kaplan, Jared and McCandlish, Sam and Henighan, Tom and Brown, Tom B and Chess, Benjamin and Child, Rewon and Gray, Scott and Radford, Alec and Wu, Jeffrey and Amodei, Dario},
  journal={arXiv preprint arXiv:2001.08361},
  year={2020}
}

\end{document}